\documentclass[twoside,11pt]{article}

\usepackage[abbrvbib,preprint]{jmlr2e}
\usepackage{xcolor}  
\usepackage{makecell}
\usepackage{dsfont}
\usepackage{diagbox}

\usepackage{adjustbox}

\usepackage{tikz}
\usetikzlibrary{calc, shapes, backgrounds}
\usepackage{amsmath, amssymb}

\newtheorem{rem}{Remark}
\newtheorem{prop}{Proposition}

\usepackage{lastpage}
\jmlrheading{24}{2023}{1-\pageref{LastPage}}{8/22; Revised
3/23}{7/23}{22-0902}{Christoph Jansen, Malte Nalenz, Georg Schollmeyer and Thomas Augustin}

\ShortHeadings{Comparing classifiers by generalized stochastic dominance}{Jansen, Nalenz, Schollmeyer and Augustin}
\firstpageno{1}

\begin{document}
\hypersetup{hidelinks}

\title{Statistical Comparisons of Classifiers by\\ Generalized Stochastic Dominance}

\author{\name Christoph Jansen\thanks{Corresponding author} \email  christoph.jansen@stat.uni-muenchen.de\\
  \addr Department of Statistics\\
  Ludwig-Maximilians-Universität\\
  Ludwigstr.~33, 80539  Munich, Germany
  \AND
   \name Malte Nalenz \email malte.nalenz@stat.uni-muenchen.de\\
  \addr Department of Statistics\\
  Ludwig-Maximilians-Universität\\
  Ludwigstr.~33, 80539  Munich, Germany
   \AND
   \name Georg Schollmeyer \email georg.schollmeyer@stat.uni-muenchen.de\\
  \addr Department of Statistics\\
  Ludwig-Maximilians-Universität\\
  Ludwigstr.~33, 80539  Munich, Germany
  \AND
  Thomas Augustin \email thomas.augustin@stat.uni-muenchen.de\\
  \addr Department of Statistics\\
  Ludwig-Maximilians-Universität\\
 Ludwigstr.~33, 80539  Munich, Germany}

\editor{Aarti Singh}

\maketitle

\begin{abstract}%
Although being a crucial question for the development of machine learning algorithms, 
there is still no consensus on how to compare classifiers over multiple data sets with respect to several criteria. Every comparison framework is confronted with (at least) three fundamental challenges: the multiplicity of quality criteria, the multiplicity of data sets and the randomness of the selection of data sets. In this paper, we add a fresh view to the vivid debate by adopting recent developments in decision theory. Based on so-called preference systems, our framework ranks classifiers by a generalized concept of stochastic dominance, which powerfully circumvents the cumbersome, and often even self-contradictory, reliance on aggregates. Moreover, we show that generalized stochastic dominance can be operationalized by solving easy-to-handle linear programs and moreover statistically tested employing an adapted two-sample observation-randomization test. This yields indeed a powerful framework for the statistical comparison of classifiers over multiple data sets with respect to multiple quality criteria simultaneously. We illustrate and investigate our framework in a simulation study and with a set of standard benchmark data sets.\end{abstract}

\begin{keywords}
algorithm comparison, statistical test, generalized stochastic dominance, preference system, decision theory
\end{keywords}


\section{Introduction} \label{intro}
\label{sec:1}

%
With a surge of new classification algorithms, a statistically sound way to decide if a method improves on its competitors is crucial. This task has eo ipso a multi-dimensional character: one often compares several classifiers over several data sets relying on several quality criteria, with accuracy, area under the curve, and Brier score being popular choices. Depending on the specific domain, also other criteria, such as model size, interpretability, and computational demand, may be of interest (see, e.g.,~\cite{ld2007}). Additionally, following the seminal work of~\cite{demvsar2006statistical}, increasing attention has been paid to the question of \textit{statistical significance} of observed classifier rankings, understanding the investigated set of data sets as a sample from a (potentially infinite) universe of data sets.
\\[.1cm]
In the latter context, a comparatively simple case occurs when only one quality criterion of metric scale is of interest, which is additionally commensurable over the selection of data sets at hand. Then classical statistical tests on mean differences for the respective criterion may be performed, e.g., pairwise t-tests. Non-metric and/or non-commensurable criteria are usually tackled by aggregating their values over the different data sets by some real-valued quantity. For instance, by considering rank aggregates, tests in the vain of \citet{demvsar2006statistical} can then be applied when judging statistical significance.
Although seeming quite intuitive, such aggregation-based approaches also have been shown to have severe shortcomings. First, the concrete choice of an aggregation procedure may heavily influence the resulting classifier ranking. 
Moreover, even seemingly plausible aggregation procedures may show paradoxical behavior; see, e.g., \cite{benavoli2016should}, who substantially question the validity of comparisons based on rank aggregation by demonstrating that adding further classifiers can change the rank order between the initially compared ones. Naturally, this problem becomes even more severe if the comparison is additionally carried out for more than one -- potentially differently scaled -- quality criterion simultaneously (also compare the discussion in Remark~\ref{paradoxa}).
\\[.1cm]
 Moreover, relying on a formal analogy, a discussion of classifier comparison might benefit from an embedding into the framework of social choice theory, where the aggregation of multiple inputs is a well-established topic (e.g., \cite{bf2002}). Examples utilizing this embedding include \citet{ehl2012} and \cite{mptbw2015}). However, this way of proceeding typically leads to impossibility results: Following~\cite{arrow1950}, it can be shown that no aggregation rule exists that satisfies a set of weak and plausible minimal requirements. This impossibility also applies when seemingly intuitive rules are used for classifier comparison. Examples are rules in the spirit of~\cite{b1781}, which evaluate classifiers by their average ranks along the different data sets and/or different criteria,  as well as rules in the spirit of~\cite{c1785}, where classifiers are ranked by counting which performs better with respect to more quality criteria and/or on more data sets.
\\[.1cm]
In summary, these considerations suggest that aggregation should be avoided unless it can be naturally derived from the underlying problem domain. Instead, it is important to look for comparison frameworks that fully exploit the information but still allow for incomparability if the available evidence is unclear. To develop such a framework, we first specify and formalize the  problem under consideration. 
%
\subsection{Specification and Formalization of the Problem}
%

To describe the different levels at which problems can arise when comparing classifiers, we use the following notation, which will be generalized in Section~\ref{sec:3}: Let $\mathcal{C}$ denote the set of classifiers under consideration, and $\mathcal{D}$ denote a set of data sets with respect to which the classifiers are to be compared. Let further $\phi_1 , \dots , \phi_n: \mathcal{C} \times \mathcal{D} \to \mathbb{R}$ be different quality criteria to measure the goodness of classification of the different classifiers on the different data sets, estimated, e.g.,~using cross-validation on each data set. 
The structure of the problem is summarized in Table~\ref{dtable} for the situation that $\mathcal{C}=\{C_1, \dots C_q\}$ and $\mathcal{D}= \{D_1 , \dots , D_s\}$. A closer look at Table~\ref{dtable} shows three different levels of challenges when comparing classifiers:
\begin{table}[ht]
\centering
\begin{tabular}{|c|ccc|}
  \hline
 \diagbox{classifiers}{data sets} & $D_1$ & $\dots$ & $D_s$ \\[.2cm]
  \hline
$C_1$ &  $\left(\begin{array}{c}\phi_1 (C_1,D_1)\\ \vdots \\ \phi_n(C_1,D_1)\end{array}\right)$ & $\dots$& $\left(\begin{array}{c}\phi_1 (C_1,D_s)\\ \vdots \\ \phi_n(C_1,D_s)\end{array}\right)$ \\ 
$\vdots$ & $\vdots$ & $\vdots$ & $\vdots$ \\ 
$C_q$ & $\left(\begin{array}{c}\phi_1 (C_q,D_1)\\ \vdots \\ \phi_n(C_q,D_1)\end{array}\right)$ & $\dots$  & $\left(\begin{array}{c}\phi_1 (C_q,D_s)\\ \vdots \\ \phi_n(C_q,D_s)\end{array}\right)$ \\ 
   \hline
\end{tabular}
\caption{A schematic presentation of the problem of comparing different classifiers over multiple data sets with respect to multiple quality criteria simultaneously.}
\label{dtable}
\end{table}
\\[.1cm]
\textit{Level 1:} In the case of multiple quality criteria, two classifiers can generally not be trivially compared already on \textit{one single} data set. For instance, consider a situation with conflicting quality criteria such as $\phi_1 (C_1,D) > \phi_1 (C_2,D) $ but at the same time $\phi_2 (C_1,D) < \phi_2 (C_2,D)$. 
Without further assumptions, no decision between the classifiers can be made in such situations: The component-wise dominance relation is a \textit{partial order} (compare Section~\ref{subsec:2}). 
\\[.1cm]
\textit{Level 2:} Even if the problem in Level~1 can be circumvented (for instance if we indeed happen to have component-wise dominance), the rank order of classifiers that holds over one fixed data set may change or even completely reverse over \textit{another} data set. For instance, it might hold that $\phi_i (C_1,D_1) > \phi_i (C_2,D_1) $ for all $i \in \{1 , \dots ,n\}$ but there exists some $i_0 \in \{1 , \dots ,n\}$ such that $\phi_{i_0} (C_1,D_2) < \phi_{i_0} (C_2,D_2) $. This makes the comparison of classifiers a decision problem \textit{under uncertainty} about the data sets, and, of course, this uncertainty should be adequately included in any further analysis of the problem. 
\\[.1cm]
\textit{Level 3:} Since both the set of all relevant data sets and their probability distribution will in general be unknown, it is often impossible to analyze the decision problem from Level~2 in practice. Instead, one can only analyze an empirical counterpart of the problem over a sample of data sets. This means that even if one has found ways to meaningfully solve the problems of Levels~1 and 2 and thus could define a meaningful order of classifiers for the concrete sample of data sets, a different order of classifiers could occur as soon as another sample of data sets is considered. The solution of such an empirical decision problem is subject to \textit{statistical uncertainty}. It is desirable to control this statistical uncertainty by constructing a suitable statistical test. 
%
\subsection{Relevance of Classifier Comparison and Related Work} \label{relwor}
The comparison of classifiers -- or more generally learning algorithms -- over multiple data sets is a much-studied and widespread problem in machine learning research. For example, in the classical benchmark setting, the performance of e.g.~classifiers over different data sets is considered in order to identify general patterns in the ranking of methods that hold 
independently of a concrete data set. This type of classifier ranking is essential for practitioners to make an informed choice about which methods, out of the huge variety of potential methods, should be considered for a given data set and evaluated, e.g.~using cross-validation. The same applies to researchers who want to improve on \textit{the best} existing methods. Therefore, benchmarking precedes the estimation of generalization performance, as it constitutes the set of models considered for any given problem domain. Examples in this spirit include \cite{MEYER2003169,Hothorn,ehl2012,mptbw2015,BISCHL201641}, to name only a few. 
\\[.1cm]
Taking into account that also the considered benchmark suite is only a selection of data sets, the importance of detecting \textit{statistically significant differences} between classifiers is shown by a whole series of high-impact publications. A particularly influential paper is \cite{demvsar2006statistical} (as well as its generalizing follow-ups~\cite{Garca2008AnEO,GARCIA20102044}), proposing a rank-based statistical test to check if one classifier is preferable to another based on a sample of data sets. Further examples include more recent papers like \cite{benavoli2016should}, who propose improvements for post hoc testing after classifier comparison over multiple data sets, or \cite{corani2017statistical}, who propose a Bayesian testing approach and call the ``\textit{statistical comparison of learning algorithms [...] fundamental in machine learning}'', see also \cite{bcdz2017},~\cite{de2017joint},~\cite{borja} and \cite{c2020} in a very similar vain. For a general review of the statistical assessment of the performance of supervised classification algorithms, see ~\cite{santafe2015dealing}; for an implementation of the framework, see \cite{calvo2016scmamp}. Typical studies include~\cite{fernandez2014we}, where algorithms are ranked based on mean accuracy on each data set, \cite{ismail2019deep}, where time series classifiers are compared, or \cite{graczyk2010nonparametric}, where neural networks for regression are investigated.
\\[.1cm]
Moreover, the variety of quality criteria, each representing different facets of the performance quality, is also a much-noticed problem (\cite{ld2007,Yu:Kumbier:2020:PNAS}): If the concrete classification task allows expressing the performance satisfactorily by a (one-dimensional) criterion, the problem should, of course, be evaluated only by this criterion. Often, however, one wants to evaluate the performance via different criteria (for example, accuracy, interpretability, and model complexity). Then, it is highly relevant to be able to make statements about which classifier can best fulfil all quality dimensions simultaneously. Further, allowing for multiple criteria also reduces the amount of potential arbitrariness of choosing one criterion out of multiple suitable candidates and hence the \textit{researcher's degrees of freedom} \citep{simmons2016false}. \\[.1cm]
Finally, selecting classifiers with respect to multiple quality criteria is strongly connected with multi-objective optimization, where one can find and analyze the Pareto-optimal points \citep{mussel2012multi,deb2014multi,jsa2023}. Approaches trying to solve such problems by aggregates that combine and trade-off multiple quality criteria into one single number are,~e.g., discussed in~\citet{brazdil2003ranking} or \citet{marler2010weighted}. Contrarily, the generalized stochastic dominance concept used in this paper, which is closely related to the one in~\cite{jsa2018,jbas2022}, allows a simultaneous evaluation of all criteria involved, instead of retreating to only one aggregation rule. 
\\[.1cm]
Generally, addressing multidimensional comparison problems by concepts of stochastic dominance is promising, and approaches in this spirit are common in fields ranging from biometrics (e.g., \cite{davidov}) to econometrics (e.g., \cite{whang2019econometric}). While the literature on testing and/or checking algorithms for (first-order) stochastic dominance is very rich in statistical contexts (e.g., \cite{f1989,m1991,m1995,bd2003,sja2017,ro2019,Jansen2022}), machine learning research relying on stochastic dominance seems to be scarce. Exceptions include \cite{Dai:StochDomConst:AIStats:2023}, who considers optimization under stochastic dominance constraints and \cite{uai2023_all}, who investigate stochastic orders for random variables with locally varying scale of measurement.  
%
\subsection{Properties of our Approach and Overview}
%
We propose a framework that 
\begin{itemize}
    \item allows a \textit{comparison} of classifiers over multiple data sets with respect to multiple quality criteria \textit{simultaneously} (therefore accounting for Levels~1 and 2 above), and 
    \item  additionally \textit{addresses statistical uncertainty} arising from the specific selection of the set of benchmark data sets (compare Level~3) by a permutation-based statistical test. 
\end{itemize}
To achieve this, we embed the problem of classifier comparison into a decision-theoretic framework and propose a generalized notion of stochastic dominance that is
\begin{itemize}
 \item \textit{information exhaustive}, fully exploiting ordinal and partial metric information in the criteria of classification quality, and thus it is
\item typically going \textit{beyond first-order stochastic dominance} on partially ordered sets, in particular providing a more expressive ordering than a Pareto analysis, while it is 
\item still \textit{avoiding the pitfalls of aggregation}.
\end{itemize}
 Notably, the ordering power of our dominance relation can be explicitly modelled by a parameter whose increase attenuates each quality dimension to the same extent. Still, our criterion generally provides  a partial ranking of the classifiers, deliberately allowing for incomparability if there is not enough evidence or conceptual rigour for a clear distinction. 
\\[.1cm]
Deriving and discussing our framework, this paper is organized as follows: Section~\ref{subsec:2} recalls required mathematical definitions. Section~\ref{sec:3} introduces the concept of $\delta$-dominance between classification algorithms, while Section~\ref{sec:4} gives an algorithm for detecting $\delta$-dominance and discusses how to test for it if only a sample of data sets is available. Sections~\ref{simulation}~and~\ref{sec:5} demonstrate the ideas presented on simulated data and with a set of standard benchmark data sets. Section~\ref{sec:6} elaborates on some promising future research perspectives.

\section{Preliminiaries}
\label{subsec:2}
Throughout the paper, we consider \textit{binary relations} at several points, be it on the set of all quality vectors as in Equation~(\ref{r1}), on a binary relation itself as in Equation~(\ref{r2}), or on the set of all classifiers as in Definition~\ref{clor}. We, therefore, begin with a compilation of some important concepts in this context. First, recall that a binary relation $R$ on a non-empty set $M$ is a subset of the Cartesian product of the set with itself, that is $R \subseteq M \times M$. Several (potential) properties of binary relation occur in the sequel: $R \subseteq M \times M$ is called \textit{reflexive}, if $(m,m) \in R$, \textit{transitive}, if $(m_1,m_2),(m_2,m_3) \in R$ implies $(m_1,m_3) \in R$, \textit{antisymmetric}, if $(m_1,m_2),(m_2,m_1) \in R$ implies $m_1=m_2$ , and \textit{complete}, if $(m_1,m_2) \in R$ or $(m_2,m_1) \in R$ (or both) for arbitrary elements $m,m_1,m_2,m_3 \in M$. A \textit{preference relation} is a binary relation that is complete and transitive; a \textit{preorder} is a binary relation that is reflexive and transitive; a \textit{linear order} is a preference relation that is antisymmetric; a \textit{partial order} is a preorder that is antisymmetric.
\\[.1cm]
Equipped with these concepts, we can now define the central ordering structure for us, so-called \textit{preference systems}. With the help of these systems, it is possible to model ordered sets on which the scale of measurement can vary locally: The (potentially partial) ordinal part of the system is modelled by a preorder on the set itself, while the (potentially partial) metric part of the system is modelled by a preorder on the ordinal relation, which covers those parts of the set for which a strength of order can also be specified. The following Definitions~\ref{ps},~\ref{consistency}, and~\ref{granularity} have been introduced in a decision-theoretic context by \cite{jsa2018} and are also discussed in \cite{jbas2022}. As these form the basis of our generalized stochastic dominance concept, they are listed here for further reference.
\begin{definition} \label{ps}
Let $A$ be a non-empty set and let $R_1 \subseteq A \times A$ denote a preorder on $A$. Moreover, let $R_2 \subseteq R_1 \times R_1$ denote a preorder on $R_1$. Then the triplet $\mathcal{A}=[A, R_1 , R_2]$ is called a \textbf{preference system} on $A$. 
\end{definition}
%
Since the definition of a preference system does not restrict the interaction of the relations $R_1$ and $R_2$ in any way, an additional consistency criterion is introduced. Here, for a preorder
$R \subseteq M \times M$ on a set $M$, we denote by $P_R \subseteq M \times M$ its \textit{strict part} and by $I_R \subseteq M \times M$ its \textit{indifference part}, respectively defined by 
$ (m_1 , m_2) \in P_{R} \Leftrightarrow (m_1 , m_2) \in R \wedge (m_2 , m_1) \notin R,$ and
$(m_1 , m_2) \in I_{R} \Leftrightarrow (m_1 , m_2) \in R \wedge (m_2 , m_1) \in R.$ 
This leads to the following definition.
\begin{definition} \label{consistency}
The preference system $\mathcal{A}=[A, R_1 , R_2]$ is \textbf{consistent} if there exists a function $u:A \to [0,1]$ such that for all $a,b,c,d \in A$ we have:
\begin{itemize}
\item[i)] If $ (a , b) \in R_1$, then $u(a) \geq u(b)$ with $=$ iff $(a,b)\in I_{R_1}$.
\item[ii)] If $((a , b),(c,d)) \in R_2$, then $u(a) -u(b) \geq u(c)-u(d)$ with $=$ iff $((a,b),(c,d))\in I_{R_2}$.
\end{itemize}
The set of all such \textbf{representations} $u$ satisfying i) and ii) is denoted by $\mathcal{U}_{\mathcal{A}}$. 
\label{consistent}
\end{definition}
For consistent preference systems possessing $R_1$-minimal and $R_1$-maximal elements in $A$, it may be useful to consider only representations that measure the utility of consequences on the same scale. This proves of particular importance for the parameter $\delta$ in Definitions~\ref{gsd} and~\ref{clor} to have a meaningful interpretation in terms of regularization. This leads to 
\begin{definition} \label{granularity}
Let $\mathcal{A}=[A, R_1 , R_2]$ be a consistent preference system containing $a_*, a^* \in A$ such that $(a^*,a) \in R_1$ and $(a,a_*) \in R_1$ for all $a \in A$. Then 
$$\mathcal{N}_{\mathcal{A}}:= \Bigl\{u  \in \mathcal{U_{\mathcal{A}}}: u(a_*)=0 ~\wedge ~ u(a^*)=1 \Bigl\}$$
is called the \textbf{normalized representation set} of $\mathcal{A}$.\\[0.1cm]
Further, for a number $\delta \in [0,1)$, we denote by $\mathcal{N}^{\delta}_{\mathcal{A}}$ the set of all $u \in \mathcal{N}_{\mathcal{A}}$ satisfying 
 $$ u(a)-u(b) \geq \delta ~~~\wedge ~~~ u(c)-u(d) -u(e) + u(f) \geq \delta $$ 
 for all $(a,b) \in P_{R_1}$ and for all $((c,d),(e,f)) \in P_{R_2} $. We call $\mathcal{A}$ \textbf{$\delta$-consistent} if $\mathcal{N}^{\delta}_{\mathcal{A}} \neq \emptyset$.
\end{definition}
Finally, we will define the notion of an automorphism in the context of preference systems, as we need this notion later in Proposition~\ref{prop_structural} and Proposition~\ref{prop_random_auto}.
\begin{definition}
Let $\mathcal{A}=[A,R_1,R_2]$ be a preference system. A mapping $T:A\to A$ is called \textbf{automorphism} if it is bijective and if furthermore for arbitrary $a,b,c,d\in A$ we have
    \begin{align*}
    (a,b) \in R_1 &\Leftrightarrow (T(a), T(b) )\in R_1 \mbox{ and } \\
    ((a,b),(c,d)) \in R_2 &\Leftrightarrow ((T(a),T(b)),(T(c),T(d))) \in R_2.
    \end{align*}
\end{definition}
\section{Comparing Classifiers by Generalized Stochastic Dominance}\label{sec:3}
We now propose a criterion for comparing classification algorithms with respect to multiple quality measures on multiple data sets \textit{simultaneously}. Specifically, our criterion generalizes \textit{(first-order) stochastic dominance} to the more structured setting of random variables taking values in a preference system (instead of only a preordered set). To begin, recall that a variable $X$ is greater or equal than another variable $Y$ with respect to (first-order) stochastic dominance if the expectation of the variable $u \circ X$ is greater or equal than the expectation of the variable $u \circ Y$ for every real-valued representation $u$ of the underlying preorder (e.g.,~\cite{m1991}). 
Our generalization now aims at increasing the ordering power by additionally allowing the partial metric information encoded in the underlying preference system to be included in the analysis.
\\[.1cm]
In our context, the variables under consideration are functions associated with the various classifiers that assign a vector of quality values, or short \textit{quality vector}, to each possible data set. The set of all these quality vectors is then partially ordered by the component-wise greater or equal relation (see Equation~(\ref{r1}) below). However, since some of the considered quality measures may also be interpretable on a metric scale, there is even more structure. To be able to exploit this partial cardinal structure, we suitably define a preference system on the set of quality vectors (see Equation~(\ref{r2}) below). A natural generalization of stochastic dominance is then to require the expectation dominance mentioned above no longer for all representations of the component-wise partial order, but only for all representations of the constructed preference system in the sense of Definitions~\ref{consistent}~and~\ref{granularity}.  
\subsection{Generalized Stochastic Dominance}
Before turning to the construction just described in detail, we give the following central definition of generalized stochastic dominance over preference systems for arbitrary random variables. This relation can be viewed as a generalization of first-order stochastic dominance in two respects: First, as just discussed, the relation $R_2$ can also include partial metric information. Second, the parameter $\delta$ allows to explicitly model from which threshold on a difference in utility should be included in the analysis of the random variables.
\begin{definition} \label{gsd}
Let $\mathcal{A}=[A, R_1 , R_2]$ be a $\delta$-consistent preference system and let $[S,\sigma(S),\pi]$ be a probability space. Denote by
$$\mathcal{F}_{(\mathcal{A},S)}:=\Bigr\{X \in A^S: u \circ X \emph{ is }\sigma(S)\emph{-}\mathcal{B}_{\mathbb{R}}([0,1])\emph{-measurable for all } u \in \mathcal{U}_{\mathcal{A}}\Bigl\}$$
For random variables $X,Y \in \mathcal{F}_{(\mathcal{A},S)}$, we say that $X$ \textbf{$(\mathcal{A},\pi , \delta)$-dominates} $Y$, abbreviated with $X \geq_{(\mathcal{A},\pi , \delta)} Y$, whenever it holds that 
$$\mathbb{E}_{\pi}(u \circ X) \geq \mathbb{E}_{\pi}(u \circ Y)$$
 for all normalized representations $u \in \mathcal{N}^{\delta}_{\mathcal{A}}$ respecting the threshold $\delta$.
\end{definition}
\begin{rem} \label{specialcases}
Consider the situation of Definition~\ref{gsd} again. For the special case that $R_2$ is the trivial preorder and $\delta=0$, the relation $\geq_{(\mathcal{A},\pi , 0)}$ essentially reduces to classical first-order stochastic dominance on partially ordered sets. Further, for the special case of $\delta=0$ and relations $R_1$ and $R_2$ that are compatible in the sense of satisfying the axioms in~\citet[Definition 1, p. 147]{k1971} and, thus, admitting a representation that is unique up to positive linear transformations, the relation $\geq_{(\mathcal{A},\pi , 0)}$ essentially reduces to the classical principle of maximizing expected utility. Finally, again setting $\delta=0$, the relation $\geq_{(\mathcal{A},\pi , 0)}$ can be viewed as that special case of the relation $R_{\forall\forall}$ from~\citet[p. 123]{jsa2018}, where the there mentioned set of probability measures $\mathcal{M}$ is chosen to consist solely of $\pi$, that is $\mathcal{M}= \{ \pi\}$ is a singleton.
\end{rem}
\subsection{Utilizing Generalized Stochastic Dominance for Comparing Classifiers} \label{gsdclass}
 As indicated at the beginning of the section, we now show how the relation $\geq_{(\mathcal{A},\pi , \delta)}$ can be utilized to compare classification algorithms with respect to multiple quality measures on multiple data sets simultaneously. This requires some additional notation. Let 
 \begin{itemize}
     \item $\mathcal{D}$ denote the set of all \textit{data sets} that are relevant for the classification task in question,
     \item $\mathcal{C}$ denote the set of all \textit{classifiers} that intend to classify the data sets from $\mathcal{D}$, 
     \item $\phi_i: \mathcal{C} \times \mathcal{D} \to Q_i$ denote a \textit{criterion of classification quality} for every $i \in \{1, \dots , n\}$,
     \item $\phi:=(\phi_1 , \dots , \phi_n):  \mathcal{D} \times \mathcal{C} \to \mathcal{Q}$, where $\mathcal{Q}:= Q_1 \times \dots \times Q_n$ is the set of \textit{quality vectors}.\footnote{Since $\mathcal{D}$ may or may not contain labels, it is not necessary to distinguish between different types of classification tasks (such as, e.g.,  supervised or unsupervised) within the proposed framework.}
 \end{itemize}
Specifically, for a data set $D \in \mathcal{D}$ and a classifier $C \in \mathcal{C}$, the (not necessarily numerical) \textit{reward} $\phi_i(C,D)$ is interpreted as the quality of the classifier $C$ for data set $D$ with respect to the classification quality criterion $\phi_i$. Importantly, note that the different reward sets $Q_1 , \dots , Q_n$ \textit{are not} assumed to be of the same scale of measurement. In particular, this implies that some of the sets are of ordinal scale (i.e.~are equipped with \textit{a preference order but no metric}), while others allow for a metric interpretation (i.e.~are equipped with \textit{both a preference order and a metric}). However, all of them are assumed to be of \textit{at least} ordinal scale and to possess minimal and maximal elements. For every $i \in \{1 ,\dots , n\}$, the preference order of the space $Q_i$ will be denoted by $\geq_i$. Note already now that these assumptions directly imply that the set $\mathcal{Q}$ possesses minimal and maximal elements w.r.t.~$R_1$ from Equation~(\ref{r1}), ensuring the normalized representation set (see Definition~\ref{granularity}) of the preference system $\mathbb{C}$ from Equation~(\ref{cps}) to be well-defined.
\\[.1cm]
Without loss of generality, we assume the quality criteria $(\phi_1 , \dots , \phi_n)$ to be arranged such that there exists $k \in \{1 ,\dots , n\}$  for which the sets $Q_1 , \dots , Q_k$ are of metric scale, equipped with metrics $d_i: Q_i \times Q_i \to \mathbb{R}$, $i=1 , \dots , k$, respectively, whereas the remaining sets are of ordinal scale not allowing for any meaningful metric interpretation. We then define a preference system on the set $\mathcal{Q}$ of all quality vectors by setting
\begin{equation} \label{r1}
    R_1 := \Bigr\{(q,p)\in \mathcal{Q} \times \mathcal{Q}: q_i  \geq_i p_i \emph{\emph{~~for all~~}}  i =1, \dots ,n \Bigl\}
\end{equation}
\begin{equation} \label{r2}
    R_2 := \Bigr\{((q,p),(r,s))\in R_1 \times R_1: d_i(q_i, p_i)  \geq d_i(r_i,s_i) \emph{\emph{~~for all~~}} i =1, \dots ,k \Bigl\}
\end{equation}
We denote the preference system which is composed of the set $\mathcal{Q}$ and the two relations just defined by $\mathbb{C}$, i.e., we have that
\begin{equation} \label{cps}
    \mathbb{C}=[\mathcal{Q},R_1, R_2]. 
\end{equation}
The two relations $R_1$ and $R_2$ can be given the following natural interpretation. Here it is important to note that the relations $R_1$ and $R_2$ do not (directly) order the classifiers themselves, but rather the quality vectors they produce on the different datasets.\\[.25cm]
\textit{Interpretation of $R_1$:} Assume we have $D \in \mathcal{D}$ and $C_i,C_j \in \mathcal{C}$ such that $\phi(C_i,D)=q$ and $\phi(C_i,D)=p$. Then $(q,p)\in R_1$ means that classifier $C_i$ has at least as high quality as classifier $C_j$ for \textit{every} considered quality measure, when evaluated on data set $D$.\\[.25cm]
\textit{Interpretation of $R_2$:} Assume we have $D \in \mathcal{D}$ and $C_i,C_j, C_k,C_l \in \mathcal{C}$ such that $\phi(C_i,D)=q$ and $\phi(C_i,D)=p$ and $\phi(C_k,D)=r$ and $\phi(C_l,D)=s$. Then $((q,p),(r,s))\in R_2$ means that, when evaluated on data set $D$, the dominance of $C_i$ over $C_j$ is \textit{at least as strong} as the dominance of $C_k$ over $C_l$. This is due to the fact that there is component-wise dominance in both cases and, additionally, the quality differences of $C_i$ and $C_j$ are at least as high as the quality differences of $C_k$ and $C_l$ for those measures that allow for a metric interpretation.\\[.25cm]
To take the final step of transferring the dominance criterion from Definition~\ref{gsd} to the comparison of classifiers, we still need to be clear about the random component in this context. This is obviously the randomness over the data sets since we are after all concerned with the expected classification quality. So, if now $[\mathcal{D},\sigma(\mathcal{D}), \pi]$ is a suitable probability space, we can use $(\mathbb{C}, \delta)$-dominance to compare classifiers with respect to all quality criteria simultaneously. To stress the crucial role of the concept, this special case deserves a separate definition for further reference. 
\begin{definition}
\label{clor}
Assume $\mathbb{C}$ to be $\delta$-consistent. For $C_i ,C_j \in \mathcal{C}$, with $\mathcal{C}$ chosen such that
$\{\phi(C,\cdot): C \in \mathcal{C} \}\subseteq \mathcal{F}_{(\mathbb{C},\mathcal{D})}$,
we say that $C_i$ $\delta$\textbf{-dominates} $C_j$, abbreviated with $C_i \succsim_{\delta} C_j$, whenever it holds that $$\phi(C_i,\cdot)  \geq_{(\mathbb{C},\pi , \delta)} \phi(C_j,\cdot)$$
In other words, it holds that $C_i \succsim_{\delta} C_j$ whenever $$\mathbb{E}_{\pi}(u \circ \phi(C_i,\cdot)) \geq \mathbb{E}_{\pi}(u \circ \phi(C_j,\cdot))$$ 
for all normalized representations $u \in \mathcal{N}^{\delta}_{\mathbb{C}}$ respecting the threshold $\delta$.
\end{definition}
\begin{rem}
Since $\delta$-dominance is a special case of $(\mathcal{A},\pi , \delta)$-dominance as introduced in Definition~\ref{gsd}, Remark~\ref{specialcases} applies here as well. In particular, if $\delta =0$ and $R_2$ is the trivial preorder, the relation $ \succsim_{\delta}$ coincides with first-order stochastic dominance. However, in the more interesting case that $R_2$ is nontrivial, the relation $ \succsim_{\delta}$ provides stronger ordering power than classical first-order stochastic dominance since it can also fully exploit the available partial cardinal information.
\end{rem}
From a decision-theoretic point of view, the threshold parameter $\delta$ can be motivated by the concept of just noticeable differences discussed in the seminal work of \cite{l1956}: It quantifies the minimal utility difference the decision maker can notice/finds relevant given utility is measured on a $[0,1]$-scale. Translated to the context of comparing classifiers, it rather can be seen as a regularization device: If some of the classifiers remain incomparable for a threshold of $\delta=0$, then increasing $\delta$ provides the opportunity to strengthen the ordering power of the dominance relation while attenuating the influence of \textit{all} quality measures used to the \textit{same degree}. This proves particularly useful for the statistical test for $\delta$-dominance discussed in Section~\ref{gsdtest}: Already very small values for $\delta$ can cause a remarkable improvement of the power of the respective test, although the basic order is only marginally changed (see also the discussion in Remark~\ref{imppoints}~and Footnote~\ref{fdelta}). 
\subsection{Some Useful Properties of the $\delta$-Dominance Relation}\label{usefulprop}
The following proposition lists some important properties of the binary relation $\succsim_{\delta}$ just introduced. Despite their elementary character, some of these properties will play an important role when applying the concepts in Sections~\ref{simulation}~and~\ref{sec:5}. 
\begin{prop}\label{prop_structural}
Consider the same situation as in Definition~\ref{clor}. The following holds:
\begin{itemize}
    \item[i)] For every $\xi \in [0, \delta]$, the relation $\succsim_{\xi}$ defines a preorder on $\mathcal{C}$.
    \item[ii)] The relations are nested with increasing $\delta$, i.e., we have $\succsim_{\xi_1} \subseteq  \succsim_{\xi_2}$ for $\xi_1 \leq \xi_2 \in [0,\delta]$.
    \item[iii)] Let $T:A\rightarrow A$ be an automorphism w.r.t.~$\mathbb{C}$. Then we have that $C_i\succsim_\delta C_j$ if and only if $C_i^T\succsim_\delta C_j^T$, where, for $p\in \{i,j\}$, $C_p^T$ represents classifier $C_p$, but evaluated not in the space $A$, but in the space $T[A]$, i.e., where $\phi(C_p,\cdot)$ is replaced by $T\circ \phi(C_p,\cdot)$.
\end{itemize}
\end{prop}

\begin{proof} i) Reflexivity is trivially true. To verify transitivity, assume that $C_i \succsim_{\xi} C_j$ and $C_j \succsim_{\xi} C_k$. Choose $u \in \mathcal{N}^{\xi}_{\mathbb{C}}$ arbitrarily (this is always possible, since $\delta$-consistency obviously implies $\xi$-consistency). Then, by assumption and definition, it holds that 
$$\mathbb{E}_{\pi}(u \circ \phi(C_i,\cdot)) \geq \mathbb{E}_{\pi}(u \circ \phi(C_j,\cdot))~~\emph{\emph{and}}~~\mathbb{E}_{\pi}(u \circ \phi(C_j,\cdot)) \geq \mathbb{E}_{\pi}(u \circ \phi(C_k,\cdot))$$ directly implying
$$\mathbb{E}_{\pi}(u \circ \phi(C_i,\cdot)) \geq \mathbb{E}_{\pi}(u \circ \phi(C_k,\cdot)).$$
As $u$ was chosen arbitrarily, this implies $C_i \succsim_{\xi} C_k$.
\\[.1cm]
ii) Assume it holds $\xi_1 \leq \xi_2 \in [0,\delta]$. By definition and $\xi$-consistency for all $\xi \in [0,\delta]$ (see i)), this implies $\emptyset \neq\mathcal{N}^{\xi_2}_{\mathbb{C}} \subseteq \mathcal{N}^{\xi_1}_{\mathbb{C}}$. Assume it holds that $C_i \succsim_{\xi_1} C_j$. By definition, this implies
$$\mathbb{E}_{\pi}(u \circ \phi(C_i,\cdot)) \geq \mathbb{E}_{\pi}(u \circ \phi(C_j,\cdot))$$
for all $u \in \mathcal{N}^{\xi_1}_{\mathbb{C}}$ and, due to the superset relation, also for all $u \in \mathcal{N}^{\xi_2}_{\mathbb{C}}$. Thus $C_i \succsim_{\xi_2} C_j$.
\\[.1cm]
iii) Because of $\{u\mid u \in \mathcal{N}^{\delta}_{\mathbb{C}}\} =\{ u\circ T\mid u \in \mathcal{N}^{\delta}_{\mathbb{C}}\}$ we have that $$\forall u \in \mathcal{N}^{\delta}_{\mathbb{C}}:  \mathbb{E}_{\pi}(u \circ \phi(C_i,\cdot)) \geq \mathbb{E}_{\pi}(u \circ \phi(C_j,\cdot)) $$ is equivalent to $$ \forall u \in \mathcal{N}^{\delta}_{\mathbb{C}}:  \mathbb{E}_{\pi}(u \circ T \circ \phi(C_i,\cdot)) \geq \mathbb{E}_{\pi}(u \circ T \circ \phi(C_j,\cdot)),$$ which shows the claim.
\end{proof}
\begin{rem} \label{paradoxa}
\cite{benavoli2016should} convincingly questioned the idea of comparing classifiers by comparing their average ranks over multiple data sets, where the ranks are computed with respect to some single quality measure $\phi$. The main problem of such an approach is that the comparison of two classifiers may depend on other classifiers that are irrelevant to the problem under consideration: Specifically, if only two classifiers $C_1$ and $C_2$ are considered, $C_1$ may get a higher average rank than $C_2$, but this relation is exactly reversed if a third classifier $C_3$ is considered, although the quality values of $C_1$ and $C_2$ are not changed. A simple example of such a situation is given in Table~\ref{comp1}.
%
\begin{table}[ht]
			\centering
			\begin{minipage}{0.45\textwidth}
			\centering
			\begin{tabular}{|c|c|c|c|c|c|}	
				\hline
				$\phi(C_i,D_j)$ & $D_1$ & $D_2$ & $D_3$ & $D_4$ & $D_5$ \\
				\hline
				$C_1$ & 0.8 & 0.8 & 0.8 & 0.6 & 0.6 \\
                $C_2$ & 0.6 & 0.6 & 0.6 & 0.8 & 0.8 \\
				\hline
			\end{tabular}
			\end{minipage}\hfill
			\begin{minipage}{0.45\textwidth}
			\centering
				\begin{tabular}{|c|c|c|c|c|c|}	
				\hline
				$\phi(C_i,D_j)$ & $D_1$ & $D_2$ & $D_3$ & $D_4$ & $D_5$ \\
				\hline
				$C_1$ & 0.8 & 0.8 & 0.8 & 0.6 & 0.6 \\
                $C_2$ & 0.6 & 0.6 & 0.6 & 0.8 & 0.8 \\
                $C_3$ & 0.9 & 0.9 & 0.9 & 0.7 & 0.7 \\
				\hline
			\end{tabular}
			\end{minipage}
			\caption{On the left, $C_1$ receives a rank sum of $8$, dominating $C_2$ with a rank sum of $7$. However, adding a third classifier $C_3$ (right table) dominating $C_1$ and $C_2$ on $D_1, D_2$ and $D_3$ and lying between $C_2$ and $C_1$ for $D_4$ and $D_5$, gives $C_2$ and $C_1$ rank sums of $9$ and $8$, respectively. The ordering of $C_1$ and $C_2$ is reversed.}
			\label{comp1}
			\end{table}
\\[.1cm]
In the case where also multiple quality criteria $\phi_1 , \dots , \phi_5$ are considered, a similar situation may already occur on one specific \textit{fixed data set $D$}. For a simple example, one can reinterpret the columns in Table~\ref{comp1} as the quality values $\phi_k(C_i,D)$ of the respective classifier with respect to the respective quality criteria on the fixed data set $D$: Comparing the average ranks of the classifiers across the quality criteria gives again reversed rankings of $C_1$ and $C_2$ for the tables on the left and on the right.  Obviously, for the case of multiple classifiers and multiple quality criteria, both problems may occur at the same time, thereby even increasing the problem. Note that this fact is well-known in social choice theory: the Borda rule from voting theory (see Section~\ref{intro}) does not satisfy Arrow's axiom of independence of irrelevant alternatives (see, e.g.,~\cite{bf2002}). 
\end{rem}
Generally, it seems that any method that uses ranks (and also any cardinal relative criterion like that used in \cite{webb2000} and discussed in \cite{demvsar2006statistical}) is akin to violating the independence of irrelevant alternatives. Note that in \cite{demvsar2006statistical} the rationale behind computing ranks is to make the quality values for different data sets commensurable. This implicitly assumes that beforehand the quality values for different data sets cannot be compared at all. If, at the same time, one also does not want to compare on one data set at least the values/ranks of different classifiers to avoid violating the independence of irrelevant alternatives, then one effectively says that one quality value of one classifier for one data set cannot be compared to any other quality value at all. This obviously will lead to an unsolvable undertaking. In our approach to ranking classifiers, we do not use ranks at all and instead demand that the quality values for different data sets are commensurable or have been made commensurable beforehand. Despite the fact that there are many decision-theoretic problems where it is difficult to specify an appropriate loss or utility function (e.g.,~\cite{Cui2021Individualized,jakubczyk2020elicitation,jsa2018}), we strongly think that a specification ensuring commensurability along the data sets is possible in our case if only all relevant details are specified to a sufficient extent.\footnote{If the latter is not the case, combining the results from~\cite{Cui2021Individualized} with our framework seems to be a promising avenue for future research.}
\\[.1cm]
Against this background, one major advantage of comparing classifiers with respect to our dominance relation $\succsim_{\delta}$ instead of applying rank-based approaches, is that, in fact, independence of irrelevant alternatives is guaranteed: 
If it holds that $C_i \succsim_{\delta} C_j$, then this statement is independent of how the space $\mathcal{C}\setminus \{C_i,C_j\}$ looks like, i.e.~of how many classifiers are considered   besides $C_i$ and $C_j$. In this way, our dominance relation circumvents one major issue recently raised in the context of rank-based comparisons.
\\[.1cm]
Another nice structural property of the relation $\succsim_{\delta}$ is that iii) in Proposition~\ref{prop_structural} is still valid for a random automorphism, as long as this random automrphism is applied independently of the process that generates the data sets. We will concretize this property in the following
\begin{prop}\label{prop_random_auto}
Let $I$ be an index set, let $\{T_z\mid z \in I\}$ be a family of automorphisms w.r.t.~$\mathbb{C}$ and let $Z:\Omega\rightarrow I$ be an indexing random variable with law $P$ that is independent of the process that generates the data sets $D$. Let, for the moment, $\mathbb{E}$ be a shorthand notation for the expectation w.r.t.~the product law $\pi \otimes{P}$. Let now $T(\omega):= T_{Z(\omega)}$ be a random automorphism. Furthermore, assume that for every $u \in \mathcal{N}_{\mathbb{C}}^\delta$ the conditional expectations $\mathbb{E}(u\circ T \circ \phi(C_i,\cdot) \mid Z=z)$ and $\mathbb{E}(u\circ T \circ \phi(C_j,\cdot)\mid Z=z)$ exist. Then we have 

\begin{align}\label{eq1}
 \forall u \in \mathcal{N}_{\mathbb{C}}^\delta:\mathbb{E}(u \circ \phi(C_i,\cdot)) &\geq  \mathbb{E}(u \circ \phi(C_j,\cdot))
 \end{align}
if and only if 
\begin{align}\label{eq2}
 \forall u \in \mathcal{N}_{\mathbb{C}}^\delta:\mathbb{E}(u \circ T\circ \phi(C_i,\cdot)) &\geq  \mathbb{E}(u \circ T \circ \phi(C_j,\cdot)).
 \end{align}
\end{prop}
\begin{proof}
Let (\ref{eq1}) hold. Then for every {\bfseries{fixed}} automorphism $T_z$ and every arbitrary $ u \in \mathcal{N}_{\mathbb{C}}^\delta$ we have 
$\mathbb{E}_\pi(u \circ T_z \circ \phi(C_i,\cdot)) \geq \mathbb{E}_\pi(u \circ T_z  \circ \phi(C_j,\cdot))$ and therefore $\mathbb{E}(u \circ T_z \circ \phi(C_i,\cdot)) \geq \mathbb{E}(u \circ T_z \circ \phi(C_j,\cdot))$ (compare Proposition \ref{prop_structural}). This implies

\begin{align*}
    \mathbb{E}(u \circ T \circ \phi(C_i,\cdot))&= \int \mathbb{E}(u \circ T \circ \phi(C_i,\cdot)\mid Z=z) \; d P(z)\\
    &= \int \mathbb{E}(u\circ T_z \circ \phi(C_i, \cdot)) \; d P(z) \geq \int \mathbb{E(}u\circ T_z \circ \phi(C_j, \cdot))\, d P(z)\\
    &= \mathbb{E}(u \circ T \circ \phi(C_j,\cdot))
\end{align*}
and therefore (\ref{eq2}) holds for every arbitrary  $ u \in \mathcal{N}_{\mathbb{C}}^\delta$. The implication (\ref{eq2}) $\Longrightarrow$ (\ref{eq1}) follows analogously by applying the corresponding inverse (random) automorphism. 
\end{proof}
Proposition~\ref{prop_random_auto}~has a nice implication: If the quality values are observed with some additional noise that can be described by a random automorphism, then the dominance criterion will not change. Note that especially a random intercept or a random scaling of the cardinal dimensions will not influence the notion of dominance. This particularly implies that in our simulation study (see Section~\ref{simulation}) we do not need to implement such random effects. 
\subsection{The GSD-$\delta$ Method}
As $\succsim_{\delta}$ defines a preorder on the set $\mathcal{C}$ of all considered classifiers (see Proposition~\ref{prop_structural}), it naturally induces an ordering structure on this set. The method of obtaining this ordering structure by relying on generalized stochastic dominance as the underlying relation will be referred to as \textsf{GSD}-$\delta$ in the following  (with \textsf{GSD}-$0$ abbreviated by \textsf{GSD}).
In order to make this method applicable in practice, two substantial questions have to be addressed. First, the question on how to efficiently check for $\delta$-dominance arises. Second, a test must be developed to judge if in-sample differences between classifiers are statistically significant. 
\section{Testing for Dominance} \label{sec:4}
In this section, we first establish a linear program for checking $\delta$-dominance between two classifiers if the set $\mathcal{D}$ of data sets is finite and the true probability law $\pi$ over this set is known. Taking into account the problem described in Level 3 from Section~\ref{sec:1}, i.e., the fact that both $\pi$ and the set $\mathcal{D}$ will in general be inaccessible, we then describe how to adapt this linear program to check for $\delta$-dominance in its empirical version, i.e., in the concrete sample of data sets drawn from the distribution $\pi$. Afterwards, we discuss how the optimal value of this adapted linear program can be reinterpreted as a test statistic for a statistical test for distributional equality of the two competing classifiers and discuss how to extend this test to the complete ordering structure between all considered classifiers. 
Finally, in preparation for the comparative study carried out in Section~\ref{simulation}, we briefly review the rank-based test proposed in~\cite{demvsar2006statistical} and suggest ways to extend it to more than one quality criterion.
\subsection{A Linear Program for Checking $\delta$-Dominance}\label{lpfcd}
We begin by discussing a linear program for checking $\delta$-dominance in the finite case. For that, consider again the preference system $\mathbb{C}$ as defined in Equation~(\ref{cps}), however, with the additional assumption that the sets $\mathcal{C}$ (the classifiers under consideration) and $\mathcal{D}$ (the data sets relevant for the comparison) are both finite. Without loss of generality, we can then assume $\mathcal{Q}=\{q_1 , \dots , q_d\}$ to be finite and that $q_1$ and $q_2$ are minimal and maximal elements of $\mathcal{Q}$ with respect to $R_1$, respectively.\footnote{As the sets $\mathcal{C}$ and $\mathcal{D}$ are finite, it makes no difference in what follows if we replace $\mathcal{Q}$ by the finite set $\phi(\mathcal{C} \times \mathcal{D})$. If $\phi(\mathcal{C} \times \mathcal{D})$ does not contain minimal and maximal elements, we define new vectors $q_1$ and $q_2$  containing a minimal or maximal element of $Q_i$ in every dimension $i$, respectively. By re-indexing the remaining vectors and considering the finite set $\phi(\mathcal{C} \times \mathcal{D})\cup \{q_1,q_2\}$ we are done.} Moreover, we assume $\delta \in [0,1)$ to be chosen such that $\mathbb{C}$ is $\delta$-consistent. A vector $(u_1, \dots , u_d) \in [0,1]^d$ then contains the images of a utility function $u:\mathcal{Q} \to [0,1]$ from $\mathcal{N}^{\delta}_{\mathbb{C}}$ if it satisfies the system of linear (in-)equalities given by
\begin{itemize}
    \item $u_{1}=0$ and $u_{2}=1$,
    \item $u_i=u_j $ for every pair $(q_i,q_j) \in I_{R_1}$,
    \item $u_i-u_j \geq \delta$ for every pair $(q_i,q_j) \in P_{R_1}$,
    \item $u_k-u_l= u_r -u_t $ for every pair of pairs $((q_k,q_l),(q_r,q_t)) \in I_{R_2}$
and 
\item $u_k-u_l-u_r+u_t \geq \delta$ for every pair of pairs $((q_k,q_l),(q_r,q_t)) \in P_{R_2}$.
\end{itemize}
Denote by $\nabla^{\delta}_{\mathbb{C}}$ the set of all vectors $(u_1, \dots , u_d) \in [0,1]^d$ satisfying all these (in)equalities. We then have the following proposition on how to check $\delta$-dominance.
\begin{prop} \label{prog}
Consider the same situation as described above. For $C_i,C_j \in \mathcal{C}$, we consider the linear programming problem
\begin{equation} \label{objfunc}
  \sum_{\ell=1}^{d} u_{\ell} \cdot [\pi(\phi(C_i,\cdot)^{-1}(\{q_{\ell}\}))-\pi(\phi(C_j,\cdot)^{-1}(\{q_{\ell}\}))] \longrightarrow \min_{(u_1 , \dots , u_d)\in \mathbb{R}^{d}}
\end{equation}
  with constraints  $(u_1 , \dots , u_d)\in\nabla^{\delta}_{\mathbb{C}}$. Denote by $opt_{ij}$ the optimal value of this programming problem. It then holds that
  $C_i \succsim_{\delta} C_j$ if and only if $opt_{ij} \geq 0$.
\end{prop}

\begin{proof} First, let $opt_{ij} \geq 0$. Choose $ u \in \mathcal{N}^{\delta}_{\mathbb{C}}$ arbitrarily and let $g:\mathbb{R}^d \to \mathbb{R}$ denote the objective function of the linear program. We then have
\begin{equation}
D(u):=\mathbb{E}_{\pi}(u \circ \phi(C_i,\cdot)) - \mathbb{E}_{\pi}(u \circ \phi(C_j,\cdot))=  g(u(q_1), \dots , u(q_d)) \geq 0
\end{equation}
where the equation follows by simple manipulations of the expected values and the lower bound of $0$ follows since, by definition, $(u(q_1), \dots , u(q_d))\in \nabla^{\delta}_{\mathbb{C}}$. Since $u \in \mathcal{N}^{\delta}_{\mathbb{C}}$ was chosen arbitrarily, this implies $C_i \succsim_{\delta} C_j$.
\\[.1cm]
Conversely, let $opt_{ij} < 0$. Choose $(u^*_1 , \dots , u^*_d) \in \nabla^{\delta}_{\mathbb{C}}$ to be an optimal solution yielding $opt_{ij}$ and define $u: \mathcal{Q} \to [0,1]$ by setting $u(q_i):=u_i^*$ for all $i = 1 , \dots , d$. We then have to distinguish two different cases:
\\[.1cm]
\textit{Case 1:} $\delta >0.$ One then easily verifies that $u \in \mathcal{N}^{\delta}_{\mathbb{C}}$ and
\begin{equation}
D(u)=  g(u_1^*, \dots , u_d^*)=opt_{ij} < 0
\end{equation}
Thus, $u$ is a function from $\mathcal{N}^{\delta}_{\mathbb{C}}$ with $\mathbb{E}_{\pi}(u \circ \phi(C_i,\cdot)) < \mathbb{E}_{\pi}(u \circ \phi(C_j,\cdot))$. Thus $\neg(C_i \succsim_{\delta} C_j)$.
\\[.1cm]
\textit{Case 2:} $\delta =0.$ If $u \in \mathcal{N}^{0}_{\mathbb{C}}$, then the same argument as in the first case applies. Thus, assume that $u \notin \mathcal{N}^{0}_{\mathbb{C}}$. Then, since $(u^*_1 , \dots , u^*_d) \in \nabla^{0}_{\mathbb{C}}$, we still know that $u$ is monotone but we no longer have \textit{strict} monotonicity with respect to the relations $R_1$ and $R_2$ of $\mathbb{C}$  (meaning that properties i) and ii) from Definition~\ref{consistency} are still valid but without the \textit{iff} condition). Now, choose $u^+ \in \mathcal{N}^{0}_{\mathbb{C}}$ arbitrarily (this is always possible, since we assume $0$-consistency). If $D(u^+)< 0$, then $\mathbb{E}_{\pi}(u^+ \circ \phi(C_i,\cdot)) < \mathbb{E}_{\pi}(u^+ \circ \phi(C_j,\cdot))$. This yields $\neg(C_i \succsim_{\delta} C_j)$. If $D(u^+)\geq 0$, then we have
$$0 \leq \xi:=\frac{D(u^+)}{D(u^+)-D(u)} <1$$
and we can choose $\alpha \in (\xi,1)$. One then easily verifies that $u_{\alpha}:= \alpha u + (1- \alpha) u^+ \in  \mathcal{N}^{0}_{\mathbb{C}}$ and that $\mathbb{E}_{\pi}(u_{\alpha} \circ \phi(C_i,\cdot)) < \mathbb{E}_{\pi}(u_{\alpha} \circ \phi(C_j,\cdot))$. This again yields that $\neg(C_i \succsim_{\delta} C_j)$, thereby completing the proof. 
\end{proof}
\subsection{A Statistical Test for $\delta$-Dominance} \label{gsdtest}
Typically, the setting discussed in Section~\ref{lpfcd} will be heavily idealized as actually we are in the situation described in Level 3 from Section~\ref{sec:1}: The true probability law $\pi$ on the set $\mathcal{D}$ as well as the set $\mathcal{D}$ itself will be unknown and inaccessible and, thus, the algorithm for checking $\delta$-dominance from Proposition~\ref{prog} will not be directly applicable. Instead of knowing the true components, we thus usually will have to work with an i.i.d. sample $D_1 , \dots , D_s \sim \pi$ of data sets from $\mathcal{D}$ in such cases. Accordingly, for defining an empirical version of the algorithm, i.e., an algorithm for checking $\delta$-dominance in the observed sample, we set 
$\hat{\mathcal{D}}_s:= \{D_1 , \dots , D_s\}$ 
and then consider the empirical law given by
\begin{equation}\label{el}
\hat{\pi}(\mathcal{W}):=\frac{1}{s}\cdot|\{j:j\in \{1, \dots ,s\}~\wedge~D_j \in \mathcal{W}\}|    
\end{equation}
for all $\mathcal{W} \in 2^{\hat{\mathcal{D}}_s}$. We then can simply run Proposition~\ref{prog} with $\mathcal{D}$ replaced by $\hat{\mathcal{D}}_s$ and $\pi$ replaced by $\hat{\pi}$. Of course, the result of this empirical version of Proposition~\ref{prog} is then subject to statistical uncertainty: even if the optimal value indicates $\delta$-dominance within the observed sample of data sets, this might not generalize to the true space $\mathcal{D}$. Conversely, it might also happen that there is $\delta$-dominance in the true space $\mathcal{D}$, while this dominance cannot be detected in the observed sample. 
\\[.1cm]
In order to control the probability of an erroneous conclusion, an appropriate statistical test should be carried out. A statistical test for the similar setup of classical stochastic dominance between random variables with values in partially ordered sets is discussed in \cite{sja2017} and based on the two-sample observation-randomization
test to be found, e.g., in~\citet[Chapter 6]{pg2012}. We now demonstrate how a such test can be transferred to our setting: As already emphasized, for the empirical version, classifier $C_i$ dominates classifier $C_j$ if and only if the optimal value $opt_{ij}$ of the linear program (\ref{objfunc}) is greater than or equal to zero. Therefore, it is natural to calculate $opt_{ij}$ in the observed sample and reject the null hypothesis
\begin{equation}
    H_0: C_j \succsim_{\delta} C_i
    \label{null}
\end{equation}
if this value is larger than a critical value $c$. Since the distribution of the statistic $opt_{ij}$ under the null hypothesis is difficult to handle, we use a permutation test that randomly swaps the roles of the classifiers for every data point, i.e., for every data set in the sample. In this way we can analyze the distribution of the test statistic under the most extreme hypothesis in $H_0$, i.e., the situation where the quality vectors of $C_i$ and $C_j$ are identically distributed. Then one can reject the null hypothesis if the value of $opt_{ij}$ for the actually observed data sets is larger than the $(1-\alpha)$-quantile of the values obtained under the resampling scheme.
\\[.1cm]
Importantly, note that we are actually interested in a statistical test that is only sensitive for deviations from $H_0$ in the direction of $\delta$-dominance in the sense of $C_i \succ_{\delta} C_j$. Therefore it would be desirable to take as the null hypothesis the negation of $C_i \succ_{\delta} C_j$. However, under this null hypothesis, the analysis of the distribution of $opt_{ij}$ seems to be difficult.  Additionally, at least for $R_2=\emptyset$ and $\delta=0$, which corresponds to classical first-order stochastic dominance, a consistent test seems to be unreachable,\footnote{A promising line of future research could be to reflect on whether the introduction of $\delta\neq 0$ and/or $R_2\neq \emptyset$ indeed leads to a consistent test for the now `regularized' null- and alternative hypotheses.}~cf., \citet[p.106]{whang2019econometric} and also \cite{gpp2019}. 
\\[.1cm]
The concrete procedure for evaluating the distribution of $opt_{ij}$ has the following five steps:  
\begin{itemize}
    \item[] \textit{Step 1:} Use the sampled data sets to produce two separate samples $(x_1 , \dots , x_s)$ and $(y_1 , \dots , y_s)$ from $\mathcal{Q}$, one for each classifier under consideration. Thereby, we used the notations $x_l:= \phi(C_i,D_l)$ and $y_l:= \phi(C_j,D_l)$ for all $l = 1, \dots ,s$.
    \item[] \textit{Step 2:} Take the pooled sample $z=(x_1, \dots , x_s, y_1 , \dots , y_s)$.
    \item[] \textit{Step 3:} Take all index sets $I \subseteq \{1, \dots , 2s \}$ of size $s$ and compute the optimal outcome $opt_{ij}^I$ of the linear program~(\ref{objfunc}) that would be obtained if $C_i$ would have produced the quality vectors $(z_i)_{i \in I}$ and if $C_j$ would have produced the quality vectors $(z_i)_{i \in \{1, \dots , 2s \} \setminus I}$.
    \item[] \textit{Step 4:} Sort all $opt_{ij}^I$ in increasing order.
    \item[] \textit{Step 5:} Reject $H_0$ if $opt_{ij}$ is greater than the $\lceil (1-\alpha) \cdot \binom{2s}{s}\rceil$-th value of the increasingly ordered values $opt_{ij}^I$, where $\alpha$ is the envisaged
confidence level.
\end{itemize}
If $\binom{2s}{s}$ is too large, instead of computing $opt_{ij}^I$ for all index sets $I$, one can alternatively compute $opt_{ij}^I$ only for a large enough number $N$ of randomly drawn index sets $I$.
\begin{rem}\label{imppoints}
Three important points should be added:
\begin{itemize}
\item[i)] If the statistical test described in (\ref{null}) is to be used to test the entire order structure on the set $\mathcal{C}$ instead of just a single pairwise comparison, it must be performed for $n \cdot (n-1)$ pairs. Then, it must be corrected for multiple testing to guarantee the specified global significance level.
\item[ii)] As already discussed at the end of Section~\ref{gsdclass}, the parameter $\delta$ acts as a regularizer. This becomes even clearer in the context of statistical testing: For $\delta=0$, the maximum value of the linear program (\ref{objfunc}) is exactly zero in the dominance case and strictly less than zero otherwise. Thus, in this case, it is impossible to compare the extent of dominance for two different dominance situations using our test statistic. If, on the other hand, we choose a value $\delta>0$ for the test, the maximum value of the linear program (\ref{objfunc}) in the dominance case can also assume values strictly greater than zero. In this way, different dominance situations can also be compared with each other in this case: The further the maximum value is above zero, the greater the extent of dominance. This potentially increases the power of the test, since situations can also be distinguished in which dominance is present in the resample, but a greater degree of dominance is present in the sample.
\item[iii)] As the parameter $\delta$ changes, of course, the hypotheses of the statistical test~(\ref{null}) also change. Thus, strictly speaking, a different statistical test is performed for each $\delta$. However, it is important to note here that the extreme case of distributional equality of the two competing classifiers for any $\delta$ belongs to the null hypothesis. Thus, the test from~(\ref{null}) for arbitrary choices of $\delta$ is suitable for detecting systematic differences in the distributions of the classifiers. Furthermore, it can be argued that a very small value of $\delta >0$, changes the order little to nothing compared to $\succsim_0$ (and thus the hypotheses of the associated statistical tests). However, it is shown (not least in the simulation study from Section~\ref{simulation}) that even such a very small value of $\delta >0$ can have a clearly visible positive effect on the power of the associated test.
\end{itemize}
\end{rem}
\subsection{Two Adoptions of Dem{\v{s}}ar's Test to Multiple Quality Criteria} \label{alttest}
As alternative methods to \textsf{GSD}-$\delta$, we now briefly review the rank-based test for comparing competing classifiers as proposed in~\cite{demvsar2006statistical} and suggest ways to extend it to more than one quality criterion. In the original test, classifiers are ranked on each data set based on their quality, typically estimated via cross-validation. Note that ranking is only straightforward for one criterion, with no obvious way to extend it to multiple dimensions.
\\[.1cm]
Ranks are then averaged over all data sets. The rationale is that data sets vary in difficulty and therefore ranking is a way to bring the different data sets on the same scale and avoid normality assumptions. The Friedman test can be applied to test for overall differences in mean-ranks. If significant differences are detected, post-hoc tests, such as the Nemenyi test can be used to determine which pairs of classifiers are significantly different. In the second step, some form of correction for multiple testing is required to hold the overall $\alpha$-level.
\\[.1cm]
As the test by \cite{demvsar2006statistical} only accounts for differences between classifiers with respect to \textit{one single} quality criterion, it must first be adapted to the setting with multiple quality criteria in order to allow a meaningful comparison with our approach. To reach a decision for multiple quality criteria we propose the following two intuitive heuristics:
\\[.1cm]
\textit{all-test}: Classifier $C_i$ is considered better than $C_j$ if it performs significantly better on each quality criterion.
\\[.1cm]
\textit{one-test}:  Classifier $C_i$ is considered better than classifier $C_j$ if $C_i$ performs significantly better in at least one dimension and if, additionally, in any other dimension classifier $C_j$ does not perform significantly better than classifier $C_i$.
\\[.1cm]
It should be noted that when using the all-test or one-test heuristics the $\alpha$-level of the one-dimensional tests is no longer preserved. In the case of the one-test, the true type 1 error will exceed $\alpha$ and is therefore no longer a valid $\alpha$-level test and instead becomes over-sensitive (cf. Section~\ref{simulation} and Appendix A3 for an example of this effect on simulated data). The all-test will often lead to a type 1 error much lower than $\alpha$, as all dimensions need to be significant and can be therefore considered as a very conservative $\alpha$-level test. We do not adjust the $\alpha$ level but note that the two approaches are perhaps the most intuitive ways to combine tests on several quality criteria.\\


%
\section{A Simulation Study} \label{simulation}
In this section, we perform a simulation study to compare the proposed statistical test for $\delta$-dominance from~(\ref{null}) for two different choices of $\delta$ with the adapted rank-based heuristics (the all-test and the one-test) as discussed in Section~\ref{alttest}. In addition, we compare how well the relation $\succsim_{\delta}$ can reproduce the order structure in the ground-truth of the simulation when evaluated only in the sample, and again contrast the rank-based sample orders comparatively. Further, we shed some
light on the role of $\delta$ for the performance of our test.
\subsection{Design of the Simulation Study}
Seven simulated classifiers $C_1, \dots , C_7$ with two-dimensional expected quality vectors $\theta_i \in [0,1]^2$ are compared. Each of the two dimensions is interpreted as a separate \textit{synthetical} quality criterion, making the simulation independent of the choice of concrete performance measures. Moreover, both quality criteria are assumed to be interpretable on a metric scale in the sense of (non-trivially) contributing to the relation $R_2$ from Equation~(\ref{r2}). The structure among the classifiers $C_1, \dots , C_7$ in the ground-truth is then induced by the recursive graph shown in Figure~\ref{sim_general}, where a \textit{separation parameter} $\eta$ controls the expected difference in performance. A specific example of such a recursive graph is given in Figure~\ref{sim_specific}. 
\\[.1cm]
The performances $x_{ij}$ of classifier $C_i$ on data set $D_j$, where $j=1,\cdots,s$,  are~i.i.d. drawn from a normal distribution, i.e., $x_{ij} \sim \mathcal{N}_2(\theta_i, \Sigma_{\epsilon})$, where $\Sigma_{\epsilon} = \sigma_{\epsilon}I$ and $\sigma_{\epsilon}$ is a noise term, which, together with $\eta$ controls the difficulty in unravelling the underlying dominance structure. Note that the $\theta_i$'s do not depend on the data set and therefore the difficulty is set to be the same for each data set. Due to Proposition~\ref{prop_random_auto}, the setting of varying difficulties is also implicitly covered for many relevant situations: In particular, a random intercept or a random scaling of the cardinal dimensions will not influence the notion of dominance, making the simulation setup quite general (see also the discussion at the end of Section~\ref{usefulprop}).
\\[.1cm]
Within this ground-truth, independent of the choice of $\eta >0$, there are ten pairs of classifiers between which there is component-wise expectation dominance. All other pairwise comparisons are set to be non-dominated in either direction, meaning that each classifier is preferable on one dimension. Note that under the assumption of independence and constant variances, the component-wise expectation dominance in the ground-truth also implies that
there are ten pairs of classifiers which are in relation with respect to the orders underlying the test for $\delta$-dominance, the all-test and the one-test, respectively. 
Note further that, since the test proposed by \citet{demvsar2006statistical} only applies for a one-dimensional quality criterion, we use the all-test and one-test heuristic described in Section~\ref{alttest} as our best effort to generalize the test to multiple dimensions for a meaningful comparison.
\begin{figure}[ht]
    \centering
    \begin{tikzpicture}[scale=1.2, transform shape]
    \node[shape=rectangle,draw=black, scale=0.5] (1) at (6,6) {$\theta_1 = \begin{pmatrix}1\\1\end{pmatrix}$};
    \node[shape=rectangle,draw=black, scale=0.5] (2) at (4,4) {$\theta_2 = \theta_1 - \begin{pmatrix}\eta\\2\eta\end{pmatrix}$};
    \node[shape=rectangle,draw=black, scale=0.5] (3) at (8,4) {$\theta_3 = \theta_1 - \begin{pmatrix}2\eta\\\eta\end{pmatrix}$};
    \node[shape=rectangle,draw=black, scale=0.5] (4) at (3,2) {$\theta_4 = \theta_2 - \begin{pmatrix}0.5\eta\\0.5\eta\end{pmatrix}$};
    \node[shape=rectangle,draw=black, scale=0.5] (5) at (5,2) {$\theta_5 = \theta_2 - \begin{pmatrix}0.25\eta\\\eta\end{pmatrix}$};
    \node[shape=rectangle,draw=black, scale=0.5] (6) at (7,2) {$\theta_6 = \theta_3 - \begin{pmatrix}\eta\\0.25\eta\end{pmatrix}$};
    \node[shape=rectangle,draw=black, scale=0.5] (7) at (9,2) {$\theta_7 = \theta_3 - \begin{pmatrix}0.5\eta\\0.5\eta\end{pmatrix}$};

    \path [->] (1) edge node[left] {} (2);
    \path [->] (1) edge node[left] {} (2);
    \path [->] (2) edge node[left] {} (4);
    \path [->] (2) edge node[left] {} (5);
    \path [->] (1) edge node[left] {} (3);
    \path [->] (3) edge node[left] {} (6);
    \path [->] (3) edge node[left] {} (7);
\end{tikzpicture}
    \caption{Simulation setting specifying the dominance between the simulated classifiers.}
    \label{sim_general}
\end{figure}
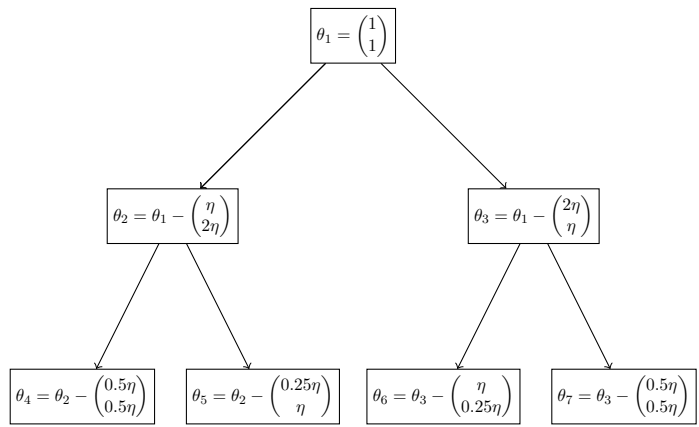
\begin{figure}[ht]
    \centering
    \begin{tikzpicture}[scale=1.2, transform shape]
    \node[shape=rectangle,draw=black, scale=0.5] (1) at (6,6) {$\theta_1 = \begin{pmatrix}1\\1\end{pmatrix}$};
    \node[shape=rectangle,draw=black, scale=0.5] (2) at (4,4) {$\theta_2 = \begin{pmatrix}0.9\\0.8\end{pmatrix}$};
    \node[shape=rectangle,draw=black, scale=0.5] (3) at (8,4) {$\theta_3 = \begin{pmatrix}0.8\\0.9\end{pmatrix}$};
    \node[shape=rectangle,draw=black, scale=0.5] (4) at (3,2) {$\theta_4 =\begin{pmatrix}0.85\\0.75\end{pmatrix}$};
    \node[shape=rectangle,draw=black, scale=0.5] (5) at (5,2) {$\theta_5 =\begin{pmatrix}0.875\\0.7\end{pmatrix}$};
    \node[shape=rectangle,draw=black, scale=0.5] (6) at (7,2) {$\theta_6 =\begin{pmatrix}0.7\\0.875\end{pmatrix}$};
    \node[shape=rectangle,draw=black, scale=0.5] (7) at (9,2) {$\theta_7 = \begin{pmatrix}0.75\\0.85\end{pmatrix}$};

    \path [->] (1) edge node[left] {} (2);
    \path [->] (1) edge node[left] {} (2);
    \path [->] (2) edge node[left] {} (4);
    \path [->] (2) edge node[left] {} (5);
    \path [->] (1) edge node[left] {} (3);
    \path [->] (3) edge node[left] {} (6);
    \path [->] (3) edge node[left] {} (7);
\end{tikzpicture}
    \caption{Example for $\Delta = 0.1$. Each entry in $\theta$ is the expected value on the corresponding quality dimension drawn from normal distribution with fixed variance and normalized to $[0,1]$.}
    \label{sim_specific}
\end{figure}
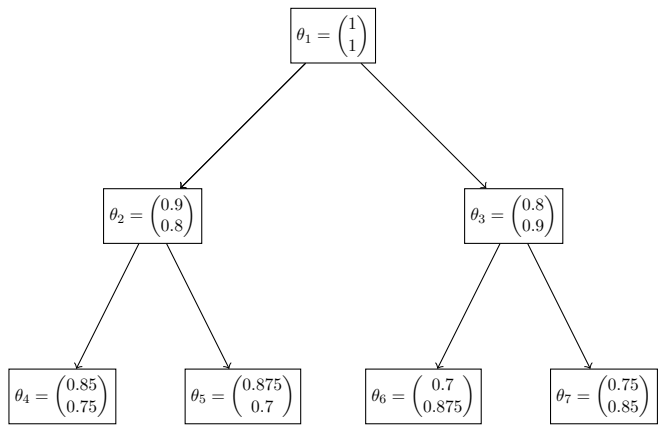
\subsection{Results} \label{results_sim}
\begin{figure}[ht]
    \centering
    \includegraphics[width = 12cm]{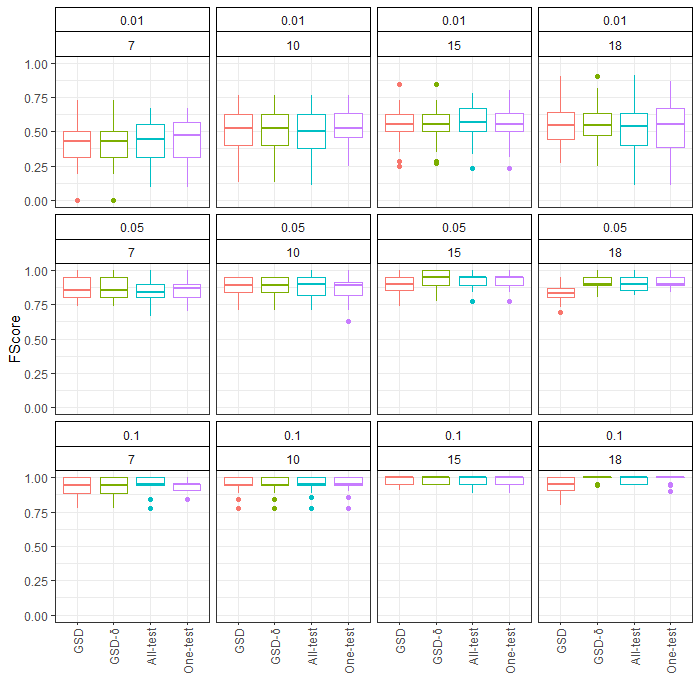}
    \caption{The figure shows the empirical distribution of the F-scores (larger is better) \textit{in the samples} (without statistical test) of the different methods along the 25 simulation runs separately for the twelve different scenarios. The F-score is computed by counting the number of TPs, FPs, and FNs in the respective \textit{sample} order compared with ground-truth and then evaluating the formula from Footnote~\ref{fscore}. }
    \label{FSs_sim}
\end{figure}

\begin{figure}[ht]
    \centering
    \includegraphics[width = 12cm]{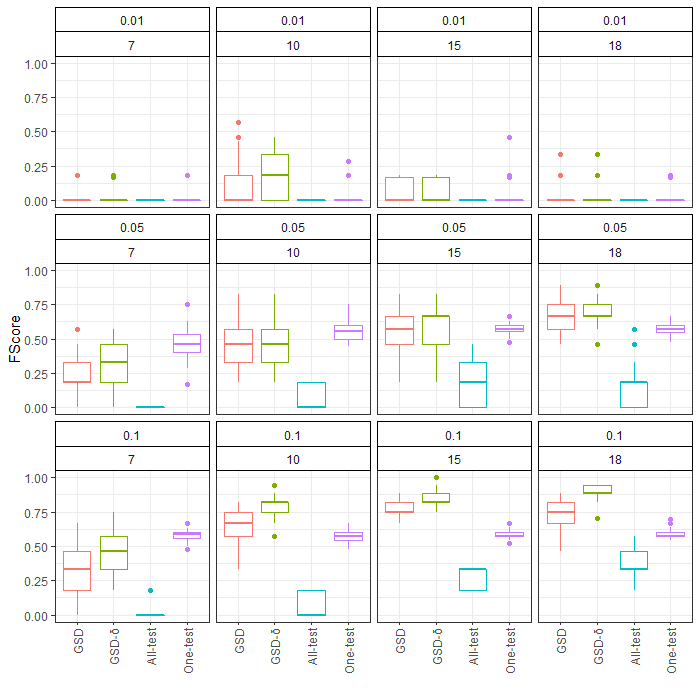}
    \caption{The figure shows the empirical distribution of the F-scores (larger is better) \textit{of the significant orders} of the different methods along the 25 runs separately for the twelve scenarios. The F-score is computed by counting the number of TPs, FPs and FNs in the respective \textit{significant order} compared with ground-truth. To account for multiple testing, the Bonferroni-correction is used in all cases.}
    \label{FSt_sim_bon}
\end{figure}

In this simulation setup, we consider a total of twelve different simulation scenarios, namely all combinations of $\eta \in \{0.01, 0.05,0.1\}$ (i.e.~varying the separation parameter) and $s \in \{7,10,15,18\}$ (i.e.~varying the number of sampled data sets). Within each simulation scenario, we carry out a total of 25 simulation runs. In each of the simulation runs, the number of resamples drawn for the corresponding resample test is chosen to increase with the number $d$ of sampled data sets. Concretely, we are interested in the following questions:
\\[.1cm]
\textit{How well does the order structure found in the sample reproduce the ground-truth?}
\\[.1cm]
To answer this question, we proceed as follows: In each simulation run, we compute the order structure of the three orders underlying the tests in the sample and compare it to the true dominance structure in the ground-truth. Specifically, we compute $\delta$-dominance in the sample by the empirical variant of the linear program from Proposition~\ref{prog} discussed earlier. For receiving the (coinciding) orders underlying the all-test and the one-test, we compute the average rank of each classifier along the data sets under each quality criterion, and then define a classifier to dominate another one whenever its average rank is superior in both quality dimensions. To measure the similarity of the orders in ground-truth and in the sample, we use the F-score, i.e., a trade-off measure between non-detected dominances (\textit{false negatives, FN}) and falsely detected dominances (\textit{false positives, FP}).\footnote{The F-Score is defined as $F=\frac{2\cdot \text{TP}}{2\cdot \text{TP}+\text{FP}+\text{FN}}$, with $\text{TP}$ the number of correctly detected dominances. An analysis under other measures (such as the Jaccard index) essentially yields the same results.\label{fscore}} 
\\[.1cm]
The results of the analyses carried out in the samples for the twelve simulation scenarios are visualized in Figure~\ref{FSs_sim}. The results show a balanced picture with regard to the different methods: All methods reproduce the order in the ground-truth about equally well. As expected, the F-score of the methods tends to improve with increasing separation parameter $\eta$ and increasing sample size $d$, with some random fluctuations which are especially visible for the lowly separated simulation scenarios.
\\[.1cm]
\textit{How well do the statistical significance tests reproduce the groundt-ruth?}
\\[.1cm]
To answer this question, under each simulation run in each scenario, we perform four different statistical tests: the $\delta$-dominance test for $\delta=0$, the $\delta$ dominance test for $\delta=10^{-5}$, the all-test, and the one-test.\footnote{The idea behind choosing an extremely small value such as $10^{-5}$ for $\delta$ in the second test, is to change the hypotheses of the test as little as possible compared to the test for $\delta=0$, but still benefit from the gain in power that a strictly positive $\delta$ brings (cf. Remark~\ref{imppoints} for further details).\label{fdelta}}
It is important to note that the one-test was included only for the sake of completeness: As already described in Section~\ref{alttest}, this test in general will not adhere to (and often drastically exceed) the specified $\alpha$-level. Thus, the comparison with significance tests at this level is of course extremely problematic. 
\\[.1cm]
To measure the similarity of the order in the ground truth and the order given by the significant edges, we again use the F-score. The used global confidence level is $\alpha=0.05$. The method used for correcting for multiple testing is the (very conservative) Bonferroni-correction for all four tests. The results of the analyses carried out at the test level for the twelve simulation scenarios are visualized in Figure~\ref{FSt_sim_bon}. 
\\[.1cm]
Some remarkable observations: Under each simulation scenario, both tests for $\delta$-dominance reveal the order structure at a global significance level of $\alpha=0.05$ at least as well as the all-test heuristic. This dominance becomes increasingly clear as the separation parameter $\delta$ and the number $d$ of simulated data sets increase. As expected, the F-score of all methods tends to improve with increasing separation parameter $\eta$ and increasing sample size $d$. Interestingly, both tests for $\delta$-dominance outperform also the one-test heuristic for separation parameters $\eta \geq 0.05$ and at least $15$ data sets, although this heuristic exceeds the given first-type error probability of $0.05$ (cf.~A3).\footnote{We note that much better trade-offs (reflected in F-score) can be found for GSD and GSD-$\delta$ if we do not enforce an overall but instead individual $\alpha$-level of 0.05 (cf.~A3 for the evaluation without Bonferroni correction, where we can see that GSD uniformly outperforms the one-test). Presumably, this could also be achieved using a more efficient multiple-testing correction strategy.}
\\[.1cm]
Furthermore, the comparison of the two tests for $\delta$-dominance for $\delta=0$ and $\delta>0$ confirms the gain in power for the latter case already theoretically indicated in Remark~\ref{imppoints} iii): The F-Score of the dominance test for $\delta>0$ exceeds the one of the dominance test for $\delta=0$ for every simulation scenario. Especially remarkable is the fact that this effect already occurs for a very small value of $\delta=10^{-5}$. Since for such a small $\delta$ the order $\succsim_{\delta}$ is presumably changed only very marginally  compared to $\succsim_0$, this suggests once more that the parameter $\delta$, in addition to its decision-theoretic interpretation, also has a pure regularization component and helps to make the hypotheses more separable.
\\[.1cm]
\textit{Summary of the results:} We have shown that the proposed statistical test for $\delta$-dominance reveals the ordering structure in the ground-truth more adequately than the all-test heuristic in each of the considered simulation scenarios. Further, we demonstrated that in the scenarios with at least medium separation ($\eta \geq 0.05$) and enough data sets available ($d \geq 15$), the tests for $\delta$-dominance also outperform the one-test heuristic, even if this heuristic does not guarantee the global $\alpha$-level. Finally, it turned out that the test for $\delta$-dominance with $\delta>0$ reproduces the order structure in all simulation scenarios at least as well as the test with $\delta=0$, even for very small choices of the parameter $\delta$.



\section{Experiments with UCI Data Sets}\label{sec:5}
We now showcase on standard benchmark data sets how the relation $\succsim_{\delta}$ and the resulting \textsf{GSD}-$\delta$ method can be used to rank classifiers based on their performance on multiple data sets with respect to multiple quality criteria. In addition, we use the statistical test proposed in Section~\ref{gsdtest} to investigate which of the orderable pairs of classifiers found in the sample may also be assumed to be statistically significant. As in Section~\ref{simulation}, we again compare our results with those obtained under the adapted rank-based tests from Section~\ref{alttest}. Finally, we examine how our results change when the analyses are based on classical stochastic dominance instead of the generalized stochastic dominance order $\succsim_{\delta}$.
\subsection{Experimental Setup} \label{expset}
For comparison, we use 16 binary classification benchmark data sets. All data sets are taken from the UCI machine learning repository \citep{Dua:2019}. The data sets strongly vary in size, dimensionality and class imbalance. For the classifier comparison, we consider the three well-established criteria \textit{accuracy}, \textit{area under the curve} and \textit{Brier score}. On each data set, 10-fold cross-validation is performed, and results are averaged for each criterion and classifier separately. Importantly, note that in the following analyses all three of these quality criteria are considered to be of metric scale and, accordingly, all equally contribute to the construction of the relation $R_2$ as most generally defined in Equation~(\ref{r2}).\footnote{An exception is of course given by Section~\ref{csd}, where a comparison with classical stochastic dominance is considered, and, thus, all three quality criteria are considered to be purely ordinal for this case.}
We compare two groups of algorithms:
\begin{itemize}
    \item For decision tree based classifiers we included  \textit{classification and regression trees} (\textsf{CART}) \citep{breiman2017classification}, \textit{random forests} (\textsf{RF}) \citep{breiman2001random}, \textit{gradient boosted trees} (\textsf{GBM}) \citep{friedman2002stochastic} and \textit{boosted decision stumps} (\textsf{BDS}) (trees with depth 2).
    \item As examples of more traditional models we included \textit{generalized linear models} (\textsf{GLM}), \textit{lasso regression} (\textsf{LASSO}) \citep{tibshirani1996regression}, \textit{elastic net} (\textsf{EN}) \citep{zou2005regularization} and \textit{ridge regression} (\textsf{RIDGE}), implemented in the \texttt{glmnet} R-package.
\end{itemize}
Generally, we expected the ensemble methods \textsf{RF} and \textsf{GBM} to dominate other methods, especially \textsf{CART}, whereas the ordering of the remaining methods is expected to be less clear. More details on data set selection, quality criteria and algorithm implementation can be found in Appendix A2.
\subsection{Results} \label{results_app}
\begin{figure}[ht]
\centering
\includegraphics[width=12cm]{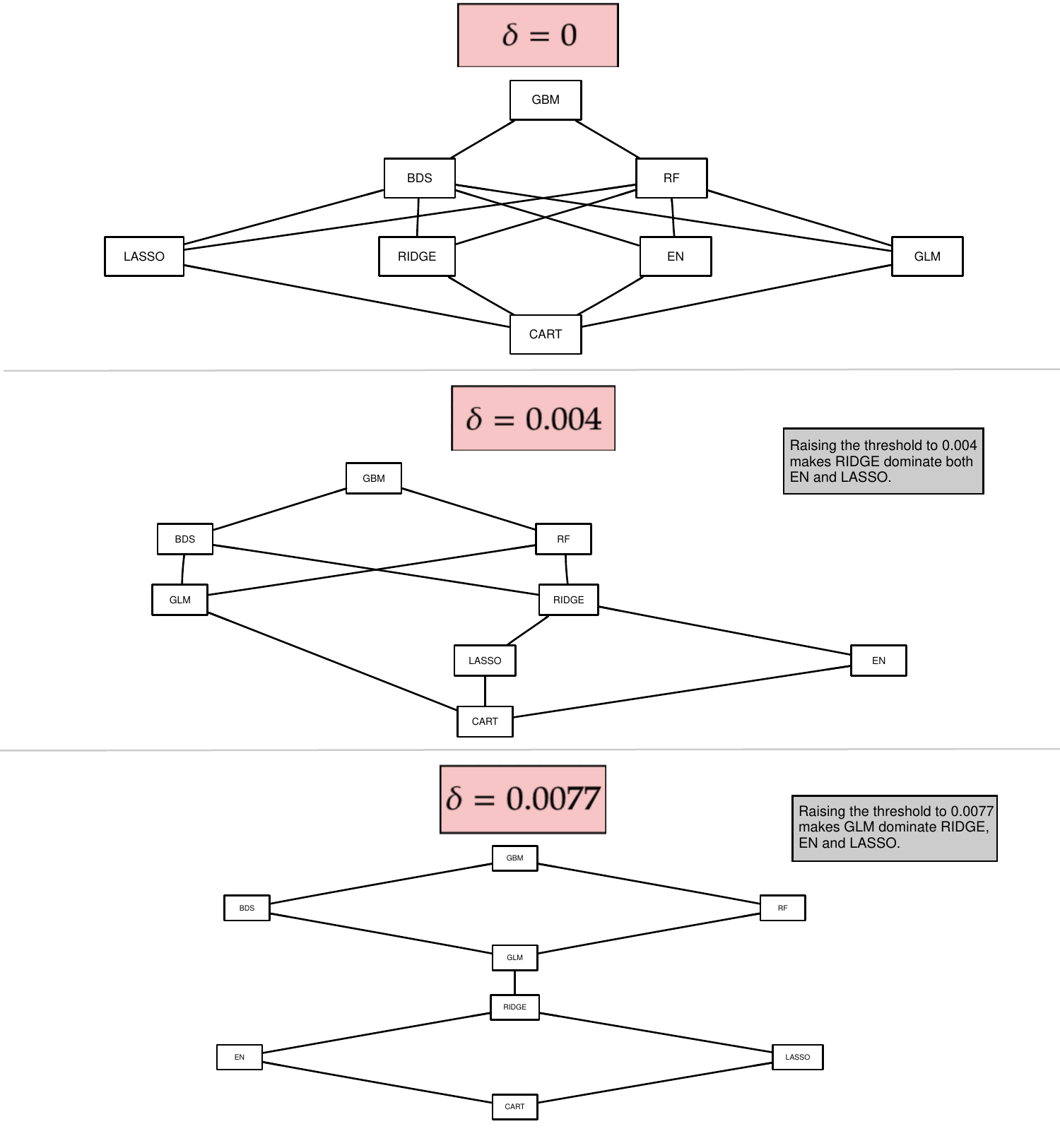}
\caption{Hasse diagrams of $\succsim_{\delta}$ in the sample for the threshold values $\delta_{min}=0$ (top),  $\delta = 0.004$ (middle)  and $\delta_{max}=0.0077$ (bottom).}
\label{fig1}
\end{figure}

In the sample of data sets just described, evaluated and visualized in the three Hasse diagrams\footnote{\textit{Hasse diagrams} are graph representations of partial orders: Whenever two nodes can be connected by a path leading top down in the graph, then the upper node dominates the lower node with respect to the considered partial order. Nodes that cannot be connected by such a top-down path are incomparable.} in Figure \ref{fig1}, the following picture emerges: even though some of the classifiers remain incomparable even for a maximum threshold of $\delta_{max}=0.0077$, concretely \textsf{BDS} and \textsf{RF} as well as \textsf{EN} and \textsf{LASSO}, a clear best classifier for this sample can be identified already for a minimum threshold of $\delta_{min}=0$, namely \textsf{GBM}. In this case also two clear second-best, but incomparable to each other, classifiers can be seen, namely \textsf{RF} and \textsf{BDS}. 
\\[.1cm]
While an analysis under $\delta_{min}$ leaves the classifier pairs 
\begin{center}
    (\textsf{GLM,RIDGE}), (\textsf{GLM, EN}), (\textsf{GLM, LASSO}), (\textsf{RIDGE, LASSO}), (\textsf{RIDGE, EN})
\end{center}
 incomparable to each other, raising the threshold to an intermediate level of $\delta = 0.004$ makes \textsf{RIDGE} dominant to both \textsf{EN} and \textsf{LASSO}.  Raising the threshold even more to $\delta_{max}$, the \textsf{GLM} classifier becomes dominant over \textsf{EN}, \textsf{LASSO} and \textsf{RIDGE}. Note that the order under threshold $\delta_{max}$ is the most structured relation we can hope for in this concrete sample: There exists no $\delta$ for which the relation $\succsim_{\delta}$ is a linear (or a preference) order. In particular, the classifiers \textsf{BDS} and \textsf{RF} as well as \textsf{EN} and \textsf{LASSO} remain incomparable in this sample no matter what threshold value is chosen.
\\[.1cm]
Next, similar to what we did in the simulation study in Section~\ref{simulation}, we perform the following three statistical tests for all pairwise comparisons: the test for $\delta$-dominance from Equation~(\ref{null}) for $\delta_{min}$ and $\delta=10^{-5}$, as well as the all-test as described in Section~\ref{alttest}. The one-test is omitted since this test will in general drastically exceed the envisaged confidence level $\alpha$, which was illustrated in the simulation study in Section~\ref{simulation}. Also note that the choice of a threshold value of $10^{-5}$ can be exactly motivated as done in Section~\ref{simulation}: Such a small value will ensure to change the hypotheses of the test as little as possible compared to the test for $\delta_{min}$, but still benefit from the gain in power that a strictly positive $\delta$ brings (compare in particular Footnote~\ref{fdelta} and Remark~\ref{imppoints}).
The results are as follows:
\\[.1cm]
\textit{Dominance tests for $\delta_{min}$:} Here we find only one of the pairwise tests to be significant on a confidence level of $\alpha=0.05$ (interpreted as a single test), namely the test of \textsf{GBM} over \textsf{BDS} (even enlarging the number of resamples from $N=1000$ to $N=10000$ does not change the situation). For all other pairwise comparison, no significant distributional difference can be identified for any confidence level smaller or equal to $0.05$. Note that under any correction procedure for multiple testing, no significant ordering structure among the classifiers can be found using this test. 
\\[.1cm]
\textit{all-test}: This test finds three pairwise comparisons of classifiers to be significant on a confidence level of $\alpha=0.05$: \textsf{BDS} over \textsf{CART}, \textsf{GBM} over \textsf{CART} and \textsf{RF} over \textsf{CART}. Again, for all other pairwise comparisons, no significant distributional difference can be identified on this level. Interestingly, the pairs of identified significant pairwise comparisons of classifiers are disjoint for this test and the resampling test for $\delta_{min}$. Note that these three pairwise comparisons of classifiers still remain significant at a global $\alpha$-level of $0.05$ under any correction procedure for multiple testing (concretely,  Bonferroni-correction was used here).
\\[.1cm]
\textit{Dominance tests for $\delta=10^{-5}$:} The results for the resample tests for all pairwise comparisons of classifiers are given in Table~\ref{rethresh}. Concretely, for every pair $(C_i,C_j)$ of classifiers, the table gives the share of resamples with test statistic strictly smaller than the test statistic in the original data, i.e., the value $\tfrac{1}{N} \sum_{I \in \mathcal{I}_N} \mathds{1}_{\{opt^I_{ij}< opt_{ij}\}}$, with $\mathcal{I}_N$ the set of resampled index sets. A line symbolizes that this share was strictly below $0.95$. 
\\[.1cm]
The table shows directly that -- interpreted as one global test on the whole ordering structure on the considered set $\mathcal{C}$ with global level $\alpha=0.05$ -- a whole series of significant pairwise comparisons emerge: First, all classifiers dominate the \textsf{CART} method significantly. Next, the \textsf{GBM} method dominates significantly all methods except \textsf{GLM} and \text{RF}. Furthermore, it can be seen that -- now interpreted as single tests at individual level $\alpha=0.05$ -- \textsf{BDS} dominates the methods \textsf{LASSO} and \textsf{RIDGE}, as well as \textsf{RF} dominates the method \textsf{EN}. Here it is important to note that the latter three pairwise comparisons are no longer significant when corrected with any procedure for multiple testing. 
\\[.1cm]
\textit{Interpretation of the results:}
Table~\ref{rethresh} shows that \textsf{GBM}, \textsf{RF} and \textsf{GLM} are not significantly different from each other and are also not significantly dominated by any other method. All other methods are, depending on the choice of $\alpha$-level and correction method, dominated by some other model.
\textsf{CART} is dominated by all other methods, including the linear model-based methods, indicating that for most data sets linear models work quite well, without strong non-linearities or interaction effects. On the other hand, single \textsf{CART} models may overfit which is a well-known issue of decision trees.
\\[.1cm]
It is also interesting to note that the regularized regression models do not perform significantly better than \textsf{GLM}. The reason might be that the regularization parameter is chosen via cross-validation, which might become unstable for smaller data sets. Also, some data sets might not include irrelevant predictors, making plain \textsf{GLM} a better choice. This could also be the reason why \textsf{GLM} is not dominated by \textsf{GBM}, as the hyper-parameters in \textsf{GBM} are not chosen via cross-validation and therefore might be suboptimal.
\textsf{GBM} as the more flexible model outperforms \textsf{BDS} for all reasonable significance levels, whereas \textsf{GBM} and \textsf{RF} are considered incomparable, which is in line with our expectations. 
\\[.1cm]
Generally speaking, using the proposed resample test for $\delta$-dominance may lead to conservative results, however, still finds more structure than the all-test heuristic, which is the only competitor we could extract from the literature. Recall again that the one-test heuristic is omitted here, as it does not hold the $\alpha$ level as discussed in Section~\ref{alttest} and, therefore, cannot be meaningfully compared to statistical tests meeting this level. As the results found by \textsf{GSD} are very trustworthy, as shown in the simulation study, we are able to make more definite statements about the performance, such as that \textsf{GBM} outperforms \textsf{LASSO} and most often \textsf{EN} and \textsf{RIDGE} regression. 
\\[.1cm]
Finally, it should be mentioned that the advantages of a small value of $\delta>0$ are also clearly shown in the concrete application: The test with $\delta=10^{-5}$ finds remarkably more structure compared to the test with $\delta_{min}$, although due to the very small value of $\delta$ the hypotheses of the underlying tests do hardly change.\footnote{As an indicator of how small this change of the underlying order actually is, one can name the fact that, restricted to the observed sample, the orders $\succsim_{0}$ and $\succsim_{0.00001}$ do actually coincide.}
\begin{table}[ht]
\centering
\begin{tabular}{rcccccccc}
  \hline
 & \textsf{BDS} & \textsf{CART} & \textsf{EN} & \textsf{GBM} & \textsf{GLM} & \textsf{LASSO} & \textsf{RF} & \textsf{RIDGE} \\ 
  \hline
\textsf{BDS} & $-$ & 1.000 & 0.976 &  $-$& $-$ & 0.967 & $-$ & 0.951 \\ 
  \textsf{CART} & $-$ &$-$  & $-$ & $-$ & $-$ & $-$ &  $-$& $-$ \\ 
  \textsf{EN} & $-$ & 0.998 & $-$ & $-$ & $-$ & $-$ & $-$ & $-$ \\ 
  \textsf{GBM} & 0.998 & 1.000 & 0.998 & $-$ & $-$ & 0.999 & $-$ & 0.997 \\ 
  \textsf{GLM} & $-$ & 1.000 & $-$ & $-$ & $-$ & $-$ & $-$ & $-$ \\ 
 \textsf{LASSO} & $-$ & 0.997 & $-$ & $-$ & $-$ & $-$ & $-$ & $-$ \\ 
  \textsf{RF}  & $-$ & 1.000 & 0.953 & $-$ & $-$ & $-$ & $-$ & $-$ \\ 
  \textsf{RIDGE} & $-$ & 0.999 & $-$ & $-$ & $-$ & $-$ & $-$ & $-$ \\ 
   \hline
\end{tabular}
\caption{Results of the resample tests with $\delta =10^{-5}$ and $N=1000$  for all binary comparisons: For every pair $(C_i,C_j)$ of classifiers, the table gives the share of resamples with test statistic strictly smaller than the test statistic in the original data. A line symbolizes that this share was strictly below $0.95$. }
\label{rethresh}
\end{table}
\subsection{Comparison with First-Order Stochastic Dominance} \label{csd}
To complete our study, we briefly compare the analysis results from Section \ref{results_app} with the results that would be obtained under an analysis under (first-order)  stochastic dominance. As discussed in Remark~\ref{specialcases} in Section~\ref{sec:3}, (first-order) stochastic dominance arises as that special case of our dominance relation $\succsim_{\delta}$ where the relation $R_2$ from the preference system $\mathbb{C}$ defined in Equation~(\ref{cps}) is the trivial preorder, and the threshold parameter $\delta$ is chosen to be $0$. Importantly, note that $R_2$ being the trivial preorder corresponds to the situation in which all the quality measures used are interpreted on a purely ordinal scale. 
\\[.1cm]
The detailed analysis results of the considered sample of data sets under classical stochastic dominance can be found in Appendix~A1. In summary, we observe that an analysis under our dominance relation $\succsim_{\delta}$ (with non-trivial $R_2$ and maybe even $\delta>0$) allows for a much more structured comparison of the competing classifiers than is the case with an analysis under classical stochastic dominance. This is due to the fact that our dominance relation also allows us to fully exploit the information of the metrically interpretable quality criteria (here, all three). In contrast, stochastic dominance considers all dimensions of the quality vectors as purely ordinal. Thus, the application example suggests that ignoring available (partial) metric information may indeed lead to ignoring relevant information about the underlying ordering structure.
\section{Summary and Concluding Remarks}\label{sec:6}
In this paper, we have developed a general framework for comparing classifiers with respect to different quality criteria on different data sets simultaneously. The basic idea of this comparison is based on a generalized version of classical multidimensional stochastic dominance, which also allows to include adequately the metric information of the quality measures used and can be regularized while attenuating the influence of the quality measures to the same extent. We have demonstrated how this dominance relation between classifiers can be detected by linear programming. Further, we showed how the optimal value of the linear program applied to a sample of data sets can be used as a statistic for statistically testing whether there is dominance between two competing classifiers. As the distribution of our test statistic is difficult to analyze, this test was performed by means of a permutation-based adapted two-sample observation-randomization test. Finally, we have illustrated the benefits of the proposed dominance concept over existing methods in a simulation study and applied it to real-world data sets comparing eight classifiers with respect to three quality criteria. 
There are several promising directions for future research:
\\[.1cm]
\textit{Incorporating classification difficulty:} At this stage, the construction of our ordering $\succsim_{\delta}$ does not incorporate differences in the difficulty of classifying different data sets.  However, as the heterogeneity of the considered space of data sets $\mathcal{D}$ increases, a co-consideration of these differences becomes more and more relevant. In principle, we believe that there would be two different ways to account for this: First, the quality vectors of the classifiers can be transformed before the dominance analysis with a loss function that depends on the data set. In this way, the challenges in comparing the difficulties would be outsourced to a pre-processing step. Of course, however, finding suitable loss functions is a research topic of its own. A second possibility is to incorporate the classification difficulty directly into the modeling of the underlying preference system. For this purpose, the framework of \textit{state-dependent} preference systems recently developed in~\cite{ja2022} would presumably be directly transferable.  
\\[.1cm]
\textit{Reducing computational complexity for special cases:} In its current form, the number of constraints of the linear program for checking $\succsim_{\delta}$-dominance given in Proposition~\ref{prog} increases with a complexity of at worst $\mathcal{O}(d^4)$, where $d$ denotes the number of possible quality vectors (or the number of attained quality vectors in the observed sample, respectively). It certainly deserves further research on how this worst-case complexity can be reduced if additional constraints on the considered preference system's metric relation $R_2$ are imposed.
\\[.1cm]    
\textit{Extension to multi-criteria decision making:} The concepts presented here need by no means be limited to the comparison of classifiers. Thinking a bit more abstractly, any algorithms could be statistically compared with respect to different performance measures simultaneously in exactly the same way, including regression and even unsupervised learning settings, as long as meaningful quality criteria can be formulated. In principle, our framework could also be applied to general multi-criteria decision problems under uncertainty. An interesting aspect is that also multi-criteria decision problems can be analyzed with respect to purely ordinal as well as metrically scaled decision rules simultaneously.

\acks{We thank three reviewers and the action editor for constructive comments that helped to improve the presentation of the paper. Financial and general support by the LMU Mentoring Program (MN and GS) and the Federal Statistical Office of Germany within the cooperation ``Machine Learning in Official Statistics'' (TA and MN) is gratefully acknowledged.}

\appendix
\section{}\label{app:theorem}
\subsection{Comparison with First-Order Stochastic Dominance}\label{a1}
\begin{figure}[h]
\centering
\includegraphics[width=11cm]{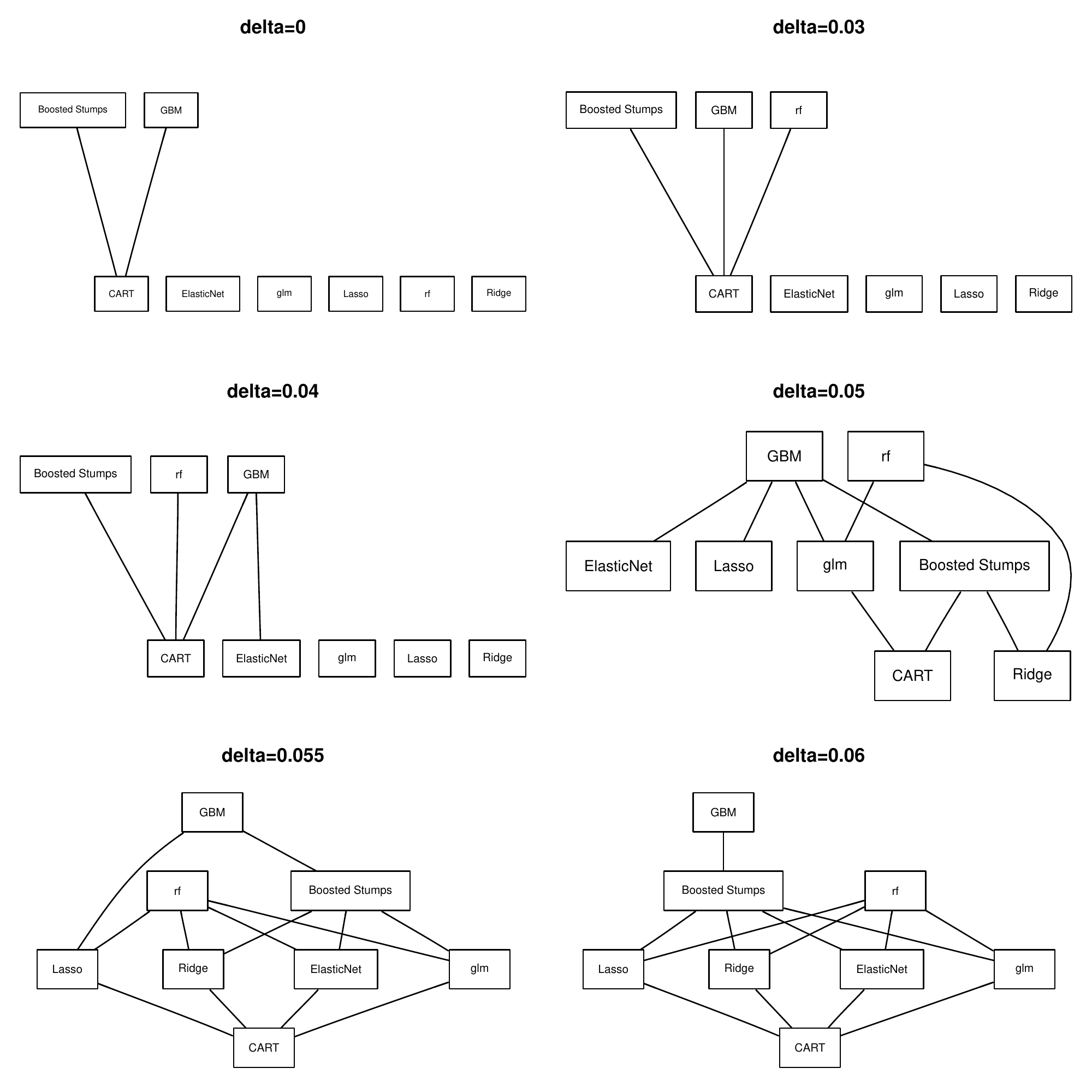}
\caption{Hasse diagrams of $\succsim_{\delta}$ for six different threshold values $\delta$ with all quality criteria treated purely ordinal. The relation $\succsim_0$ coincides with stochastic dominance.}
\label{fig2}
\end{figure}

If we analyze the same situation as described in Section \ref{expset}, however, this time by means of (first-order) stochastic dominance, we receive the results which are visualized in Figure~\ref{fig2}. The first major difference from the analysis based on our dominance relation is the ordering under a threshold of $\delta=0$: While a relatively structured picture emerged for the analysis under our dominance relation $\succsim_{0}$ already in this case (compare the top picture in Figure~\ref{fig1}), the analysis based on (first-order) stochastic dominance yields only two pairs of comparable classifiers, viz \textsf{BDS} over \textsf{CART} as well as \textsf{GBM} over \textsf{CART}. Based on stochastic dominance, no clear best and worst classifiers can be identified within this concrete sample.
\\[.1cm]
Considering the remaining five analyses under successively increasing threshold $\delta$, two aspects, in particular, should be emphasized. First, as expected, the higher the threshold value, the more comparable the classifiers. In comparison to the analysis based on our dominance relation $\succsim_{\delta}$ under increasing $\delta$, however, it is noticeable that even with threshold values that are higher by a factor of about ten, there still arise more weakly structured situations.
Second, it is striking that even when analyzed with a relatively high threshold of $\delta=0.06$, no clear best classifier can be identified: The methods \textsf{RF} and \textsf{GBM} remain incomparable here. Note that the order under threshold $\delta =0.06$ is the best we can get: There exists no $\delta$ for which $\succsim_{\delta}$ possesses a superset of comparable pairs of classifiers.
\subsection{Data Set Selection and Implementation in Section~\ref{expset}}
\textit{Data Set Selection.} The data sets used for the experiments in Section~\ref{expset} are taken from the UCI machine learning repository \citep{Dua:2019}. The selection criteria are: We only consider binary classification, but note that the method can be extended to any learning task where performance criteria of at least ordinal scale exist. We chose data sets with mostly numerical features or features with low cardinality. Only data sets with a low number of missing values are considered.
Generally, we selected data sets, such that the need for pre-processing is minimal.
\\[.25cm]
\textit{Algorithm Settings.} We briefly describe the implementation of the compared methods: Ridge, Elastic Net and Lasso Regression are fit using the R-package \citep{Rcite} \textbf{glmnet} \citep{glmnet}. The optimal $\lambda$ is determined via cross-validation. The mixing parameter in Elastic Net is set to 0.5.
GBM and Gradient boosted decision stumps are fit using the \textbf{gbm} R-package \citep{gbmcite}. Gradient boosting uses 300 trees with a learning rate of 0.02 and a maximum depth of 3. The stumps use 500 trees and a learning rate of 0.05. Random Forest is fit using the \textbf{randomForest} R-package \citep{rfcite} with default settings. For CART we use the \textbf{rpart} R-package \citep{rpartcite} with default settings. Note that the results for all algorithms could likely be improved with parameter tuning, however, we used reasonable default values for the comparison. Our aim is solely to showcase our method, not to make any definite statements about the general performance of popular machine learning methods.
\subsection{More Detailed Results}
\begin{figure}[h!]
    \centering
    \includegraphics[scale = 0.5]{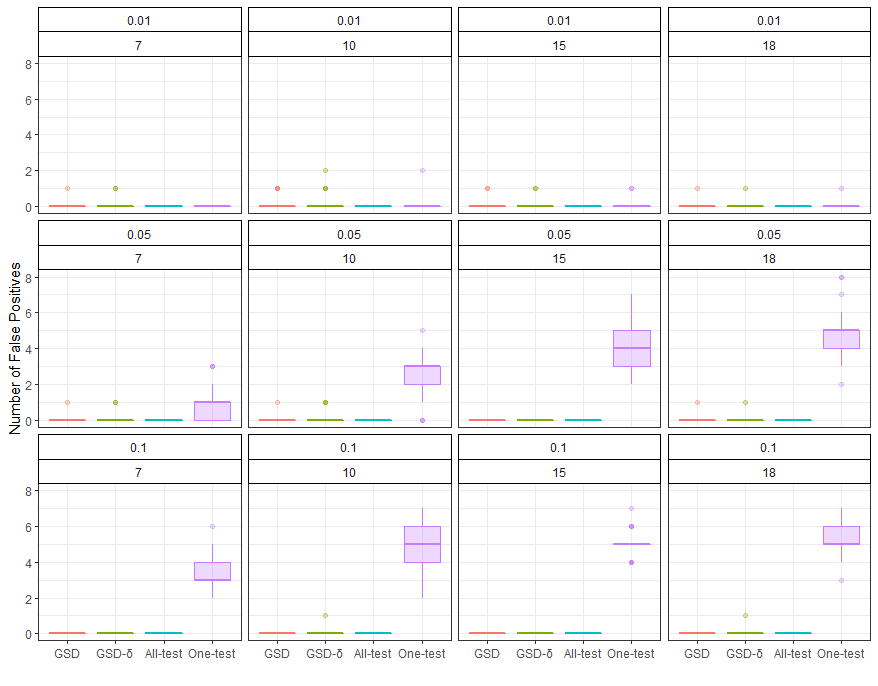}
    \caption{The figure shows the empirical distribution of the number of false positives.}
    \label{FSs_sim0}
\end{figure}
\begin{figure}[h!]
    \centering
    \includegraphics[scale = 0.5]{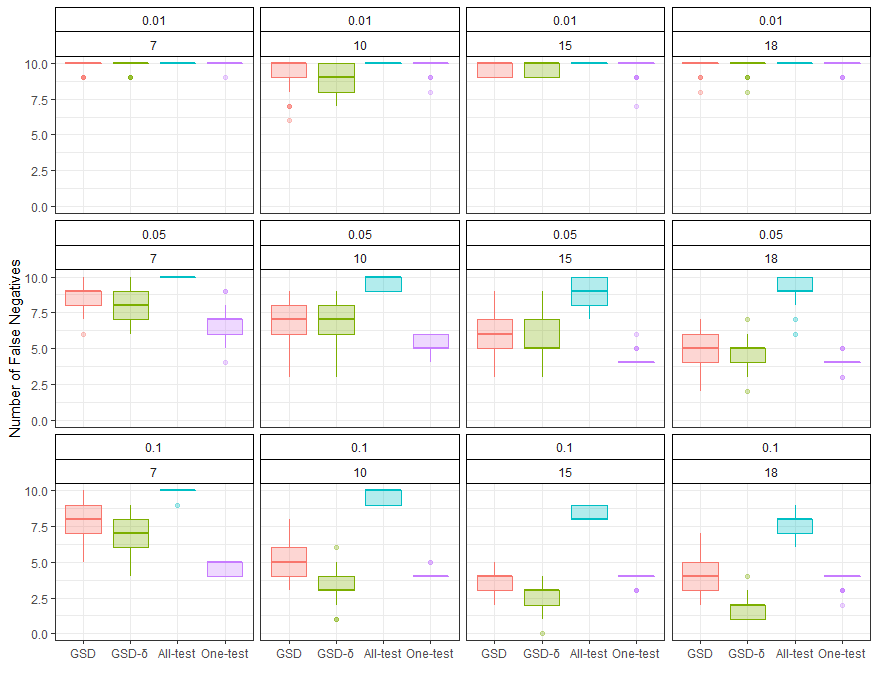}
    \caption{The figure shows the empirical distribution of the number of false negatives.}
    \label{FSs_sim1}
\end{figure}
\begin{figure}[h!]
    \centering
    \includegraphics[scale = 0.48]{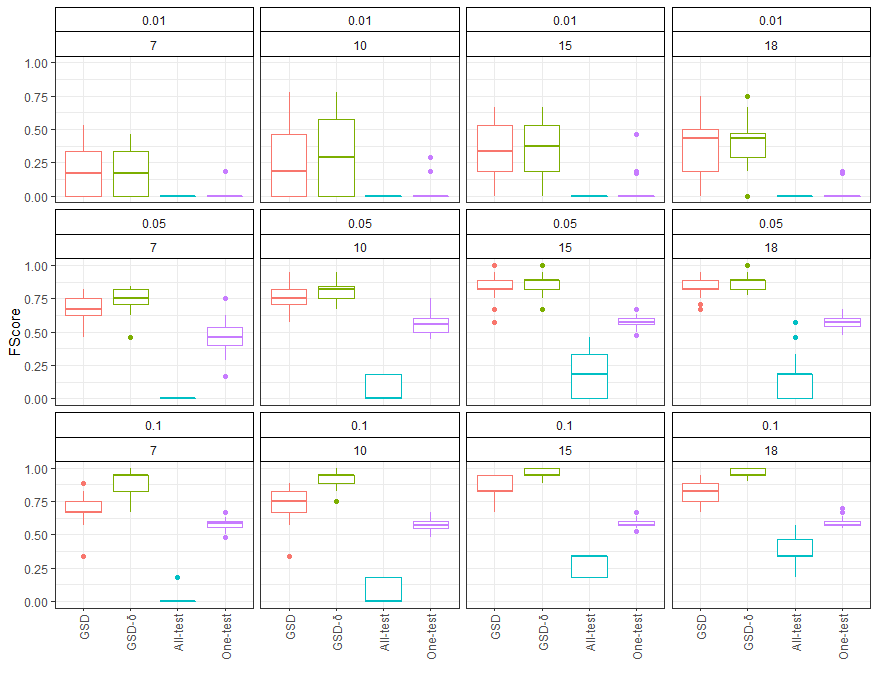}
    \caption{F-score for the different tests, without Bonferroni correction. A much better trade-off is achieved for GSD and GSD-$\delta$.}
    \label{FSs_sim2}
\end{figure}
\noindent\textit{Empirical Results.} Tables~\ref{t1},~\ref{t2}, and~\ref{t3} show the raw performance values over the 16 analyzed data sets that the classifier comparison is based on. Figures~\ref{FSs_sim0},~\ref{FSs_sim1}, and~\ref{FSs_sim2} show the empirical distributions of the number of false positives, the number of false negatives, and the F-scores for the different tests without correction for multiple testing.
\begin{table}[h!] \centering 
\begin{adjustbox}{scale = 0.45}
\begin{tabular}{@{\extracolsep{5pt}} ccccccccc} 
\\[-1.8ex]\hline 
\hline \\[-1.8ex] 
data set & Boosted Stumps & CART & ElasticNet & GBM & glm & Lasso & rf & Ridge \\ 
\hline \\[-1.8ex] 
australian & $0.937$ & $0.901$ & $0.930$ & $0.943$ & $0.929$ & $0.929$ & $0.937$ & $0.931$ \\ 
banknote & $1.000$ & $0.976$ & $1.000$ & $1.000$ & $1.000$ & $1.000$ & $1.000$ & $1.000$ \\ 
biodeg & $0.926$ & $0.847$ & $0.917$ & $0.926$ & $0.922$ & $0.917$ & $0.937$ & $0.916$ \\ 
blood\_transfusion & $0.738$ & $0.722$ & $0.751$ & $0.736$ & $0.752$ & $0.752$ & $0.670$ & $0.751$ \\ 
diabetes & $0.830$ & $0.763$ & $0.833$ & $0.838$ & $0.831$ & $0.833$ & $0.826$ & $0.832$ \\ 
haberman & $0.657$ & $0.556$ & $0.717$ & $0.697$ & $0.713$ & $0.728$ & $0.673$ & $0.720$ \\ 
heart & $0.906$ & $0.823$ & $0.904$ & $0.902$ & $0.912$ & $0.906$ & $0.904$ & $0.910$ \\ 
ILPD & $0.728$ & $0.674$ & $0.714$ & $0.732$ & $0.738$ & $0.717$ & $0.754$ & $0.719$ \\ 
Ionosphere & $0.972$ & $0.915$ & $0.910$ & $0.973$ & $0.866$ & $0.904$ & $0.980$ & $0.913$ \\ 
liver & $0.649$ & $0.587$ & $0.671$ & $0.650$ & $0.668$ & $0.662$ & $0.606$ & $0.672$ \\ 
parkinsons & $0.942$ & $0.830$ & $0.873$ & $0.957$ & $0.866$ & $0.867$ & $0.951$ & $0.853$ \\ 
pop\_failures & $0.910$ & $0.817$ & $0.943$ & $0.937$ & $0.952$ & $0.941$ & $0.923$ & $0.942$ \\ 
sonar & $0.904$ & $0.784$ & $0.831$ & $0.927$ & $0.755$ & $0.838$ & $0.946$ & $0.855$ \\ 
spambase & $0.981$ & $0.892$ & $0.952$ & $0.981$ & $0.971$ & $0.952$ & $0.986$ & $0.952$ \\ 
wbdc & $0.993$ & $0.960$ & $0.994$ & $0.992$ & $0.962$ & $0.994$ & $0.991$ & $0.993$ \\ 
wilt & $0.990$ & $0.956$ & $0.970$ & $0.989$ & $0.977$ & $0.970$ & $0.989$ & $0.963$ \\ 
\hline \\[-1.8ex] 
\end{tabular} 
\end{adjustbox}
\caption{AUC for the different methods on the 16 data sets.}
\label{t1}
\end{table} 
\begin{table}[h!] \centering 
\begin{adjustbox}{scale = 0.45}
\begin{tabular}{@{\extracolsep{5pt}} ccccccccc} 
\\[-1.8ex]\hline 
\hline \\[-1.8ex] 
data set & Boosted Stumps & CART & ElasticNet & GBM & glm & Lasso & rf & Ridge \\ 
\hline \\[-1.8ex] 
australian & $0.865$ & $0.845$ & $0.859$ & $0.864$ & $0.859$ & $0.859$ & $0.867$ & $0.859$ \\ 
banknote & $0.995$ & $0.965$ & $0.975$ & $0.989$ & $0.987$ & $0.975$ & $0.993$ & $0.977$ \\ 
biodeg & $0.870$ & $0.824$ & $0.861$ & $0.866$ & $0.865$ & $0.860$ & $0.871$ & $0.856$ \\ 
blood\_transfusion & $0.788$ & $0.784$ & $0.771$ & $0.789$ & $0.770$ & $0.771$ & $0.761$ & $0.773$ \\ 
diabetes & $0.751$ & $0.734$ & $0.769$ & $0.763$ & $0.773$ & $0.771$ & $0.768$ & $0.772$ \\ 
haberman & $0.735$ & $0.719$ & $0.735$ & $0.729$ & $0.742$ & $0.735$ & $0.725$ & $0.732$ \\ 
heart & $0.805$ & $0.786$ & $0.832$ & $0.812$ & $0.842$ & $0.839$ & $0.815$ & $0.848$ \\ 
ILPD & $0.700$ & $0.671$ & $0.710$ & $0.712$ & $0.724$ & $0.708$ & $0.705$ & $0.712$ \\ 
Ionosphere & $0.926$ & $0.875$ & $0.869$ & $0.937$ & $0.878$ & $0.872$ & $0.937$ & $0.877$ \\ 
liver & $0.583$ & $0.569$ & $0.629$ & $0.606$ & $0.615$ & $0.626$ & $0.554$ & $0.629$ \\ 
parkinsons & $0.902$ & $0.841$ & $0.876$ & $0.922$ & $0.860$ & $0.876$ & $0.907$ & $0.861$ \\ 
pop\_failures & $0.926$ & $0.928$ & $0.915$ & $0.943$ & $0.957$ & $0.915$ & $0.924$ & $0.915$ \\ 
sonar & $0.818$ & $0.755$ & $0.756$ & $0.842$ & $0.736$ & $0.737$ & $0.842$ & $0.794$ \\ 
spambase & $0.944$ & $0.893$ & $0.884$ & $0.939$ & $0.927$ & $0.884$ & $0.954$ & $0.884$ \\ 
wbdc & $0.963$ & $0.944$ & $0.961$ & $0.961$ & $0.960$ & $0.961$ & $0.963$ & $0.956$ \\ 
wilt & $0.976$ & $0.977$ & $0.943$ & $0.980$ & $0.969$ & $0.943$ & $0.982$ & $0.945$ \\ 
\hline \\[-1.8ex] 
\end{tabular} 
\end{adjustbox}
\caption{Accuracy on the 16 data sets.} 
\label{t2}
\end{table} 
\begin{table}[h!] \centering 
\begin{adjustbox}{scale = 0.45}
\begin{tabular}{@{\extracolsep{5pt}} ccccccccc} 
\\[-1.8ex]\hline 
\hline \\[-1.8ex] 
data set & Boosted Stumps & CART & ElasticNet & GBM & glm & Lasso & rf & Ridge \\ 
\hline \\[-1.8ex] 
australian & $0.095$ & $0.119$ & $0.106$ & $0.091$ & $0.101$ & $0.106$ & $0.096$ & $0.106$ \\ 
banknote & $0.011$ & $0.032$ & $0.034$ & $0.011$ & $0.009$ & $0.034$ & $0.006$ & $0.041$ \\ 
biodeg & $0.099$ & $0.141$ & $0.123$ & $0.100$ & $0.101$ & $0.123$ & $0.093$ & $0.126$ \\ 
blood\_transfusion & $0.156$ & $0.157$ & $0.159$ & $0.156$ & $0.155$ & $0.159$ & $0.183$ & $0.159$ \\ 
diabetes & $0.163$ & $0.195$ & $0.163$ & $0.158$ & $0.158$ & $0.163$ & $0.161$ & $0.163$ \\ 
haberman & $0.191$ & $0.203$ & $0.184$ & $0.187$ & $0.183$ & $0.183$ & $0.194$ & $0.184$ \\ 
heart & $0.134$ & $0.168$ & $0.132$ & $0.133$ & $0.128$ & $0.132$ & $0.131$ & $0.130$ \\ 
ILPD & $0.182$ & $0.224$ & $0.187$ & $0.179$ & $0.176$ & $0.187$ & $0.172$ & $0.187$ \\ 
Ionosphere & $0.058$ & $0.099$ & $0.114$ & $0.051$ & $0.105$ & $0.117$ & $0.050$ & $0.111$ \\ 
liver & $0.248$ & $0.278$ & $0.234$ & $0.242$ & $0.228$ & $0.235$ & $0.253$ & $0.233$ \\ 
parkinsons & $0.070$ & $0.126$ & $0.112$ & $0.061$ & $0.111$ & $0.114$ & $0.072$ & $0.117$ \\ 
pop\_failures & $0.052$ & $0.058$ & $0.061$ & $0.043$ & $0.033$ & $0.061$ & $0.055$ & $0.061$ \\ 
sonar & $0.120$ & $0.204$ & $0.181$ & $0.110$ & $0.264$ & $0.176$ & $0.124$ & $0.169$ \\ 
spambase & $0.046$ & $0.094$ & $0.112$ & $0.048$ & $0.059$ & $0.112$ & $0.039$ & $0.111$ \\ 
wbdc & $0.025$ & $0.050$ & $0.059$ & $0.028$ & $0.040$ & $0.059$ & $0.030$ & $0.062$ \\ 
wilt & $0.017$ & $0.019$ & $0.044$ & $0.015$ & $0.024$ & $0.044$ & $0.013$ & $0.044$ \\ 
\hline \\[-1.8ex] 
\end{tabular} 
\end{adjustbox}
\caption{Brier Score on the 16 data sets.} 
\label{t3}
\end{table} 
\vskip 0.2in
\bibliography{main}
\end{document}